\documentclass[accepted]{uai2022} % after acceptance, for a revised
\usepackage[american]{babel}
\usepackage{natbib} % has a nice set of citation styles and commands
\usepackage{mathtools} % amsmath with fixes and additions
\usepackage{booktabs} % commands to create good-looking tables
\usepackage{tikz} % nice language for creating drawings and diagrams

\usepackage{url}
\usepackage{algorithmic}
\usepackage{algorithm}
\usepackage{graphicx}
\usepackage{subfig}

\usepackage{cleveref}
\usepackage{hyperref}
\hypersetup{colorlinks=true, linkcolor=black, urlcolor=black, citecolor=black}

\renewcommand{\algorithmiccomment}[1]{\bgroup\hfill\scriptsize\it//~#1\egroup}
\newcommand{\algorithmicbreak}{\textbf{break}}
\newcommand{\BREAK}{\STATE \algorithmicbreak}

\newlength{\figwidth}
\newlength{\figheight}
\usepackage{amsmath,amsfonts,bm}

\def\1{\bm{1}}

\def\rvx{{\mathbf{x}}}
\def\rvy{{\mathbf{y}}}
\def\rvz{{\mathbf{z}}}

\def\vmu{{\boldsymbol{\mu}}}
\def\vtheta{{\boldsymbol{\theta}}}
\def\valpha{{\boldsymbol{\alpha}}}

\def\vSigma{{\boldsymbol{\Sigma}}}
\def\va{{\mathbf{a}}}

\def\vx{{\mathbf{x}}}
\def\vy{{\mathbf{y}}}
\def\vz{{\mathbf{z}}}
\def\vL{{\mathbf{L}}}

\def\mA{{\mathbf{A}}}
\def\mB{{\mathbf{B}}}

\def\mH{{\mathbf{H}}}
\def\mI{{\mathbf{I}}}
\def\mJ{{\mathbf{J}}}

\def\mU{{\mathbf{U}}}
\def\mV{{\mathbf{V}}}
\def\mW{{\mathbf{W}}}
\def\mX{{\mathbf{X}}}

\def\mSigma{{\boldsymbol{\Sigma}}}

\DeclareMathAlphabet{\mathsfit}{\encodingdefault}{\sfdefault}{m}{sl}
\SetMathAlphabet{\mathsfit}{bold}{\encodingdefault}{\sfdefault}{bx}{n}

\newcommand{\R}{\mathbb{R}}

\newcommand{\vphi}{\boldsymbol{\phi}}
\newcommand{\vpi}{\boldsymbol{\pi}}

\newcommand{\vsigma}{\boldsymbol{\sigma}}
\renewcommand{\vx}{\mathbf{x}}
\renewcommand{\vy}{\mathbf{y}}
\renewcommand{\vz}{\mathbf{z}}

\renewcommand{\va}{\mathbf{a}}

\renewcommand{\vec}{\text{vec}}

\newcommand{\MN}{\mathcal{MN}}
\newcommand{\N}{\mathcal{N}}
\newcommand{\diag}[1]{\text{diag}(#1)}

\renewcommand{\R}{\mathbb{R}}

\newcommand{\D}{\mathcal{D}}

\newcommand{\inv}{{-1}}

\renewcommand{\L}{\mathcal{L}}

\title{Fast Predictive Uncertainty for \\Classification with Bayesian Deep Networks}

\author[1]{\href{mailto:<marius.hobbhahn@gmail.com>?Subject=Your UAI 2022 paper}{Marius~Hobbhahn}{}}
\author[1]{Agustinus~Kristiadi}
\author[1,2]{Philipp~Hennig}
\affil[1]{%
    University of Tübingen
}
\affil[2]{%
    Max-Planck Institute for Intelligent Systems
}
  
\begin{document}
\maketitle

\begin{abstract}
In Bayesian Deep Learning, distributions over the output of classification neural networks are often approximated by first constructing a Gaussian distribution over the weights, then sampling from it to receive a distribution over the softmax outputs. This is costly. We reconsider old work (Laplace Bridge) to construct a Dirichlet approximation of this softmax output distribution, which yields an analytic map between Gaussian distributions in logit space and Dirichlet distributions (the conjugate prior to the Categorical distribution) in the output space. 
Importantly, the vanilla Laplace Bridge comes with certain limitations. We analyze those and suggest a simple solution that compares favorably to other commonly used estimates of the softmax-Gaussian integral.
We demonstrate that the resulting Dirichlet distribution has multiple advantages, in particular, more efficient computation of the uncertainty estimate and scaling to large datasets and networks like ImageNet and DenseNet. 
We further demonstrate the usefulness of this Dirichlet approximation by using it to construct a lightweight uncertainty-aware output ranking for ImageNet. 
\end{abstract}
%%%%%%%%%%%%%%%%%%%%%%%%
\section{Introduction}
\label{sec:introduction}
Quantifying the uncertainty of Neural Networks' (NNs) predictions is important in safety-critical applications such as medical-diagnosis \citep{UIinMedicine} and self-driving vehicles \citep{McAllister2017ConcretePF,AutoDrivingBayes}, but it is often limited by computational constraints. 
Architectures for classification tasks produce a probability distribution as their output, constructed by applying the softmax to the point-estimate output of the penultimate layer.
However, it has been shown that this distribution is overconfident \citep{nguyen2015deep,Hein_2019_CVPR} and thus cannot be used for predictive uncertainty quantification.
%%%%%%%%%%%%%%%%%%%%%%
\begin{figure}[t!]
     \centering

     \includegraphics[width=0.48\textwidth]{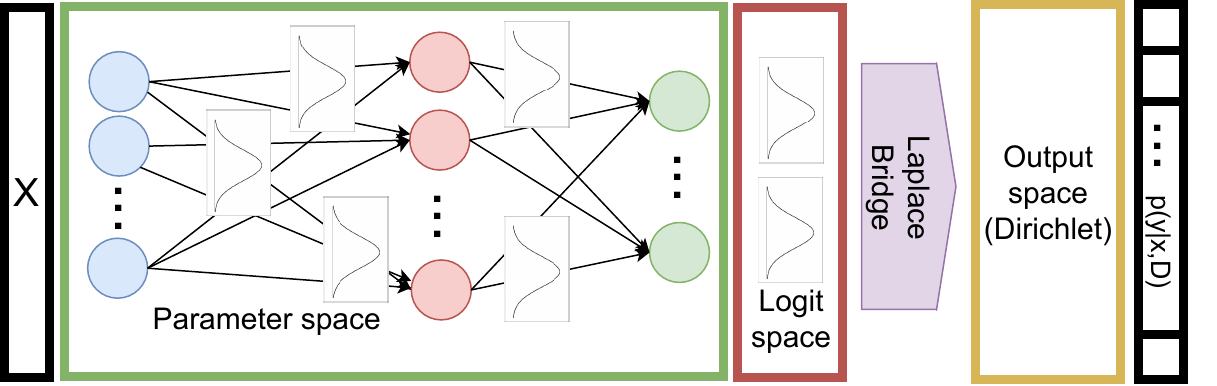}
     
     \caption{High-level sketch of the Laplace Bridge for BNNs. $p(y|x,D)$ denotes the marginalized softmax output, i.e. the mean of the Dirichlet.}
     %\vspace{-1em}
     \label{fig:LB_for_BNN_overview}
     %\vspace{-0em}
\end{figure}
%%%%%%%%%%%%%%%%%%%%
Approximate Bayesian methods provide quantified uncertainty over the NN's parameters in a tractable fashion. The commonly used Gaussian approximate posterior \citep{MacKay1992,Graves2011VB,Blundell2015WeightUI,ritter2018a} approximately induces a Gaussian distribution over the logits of a NN \citep{McKay1995NetworkBayesReview}, but the associated predictive distribution is not analytic.
It is typically approximated by Monte Carlo (MC) integration. This requires multiple samples, making prediction in Bayesian Neural Networks (BNNs) a comparably expensive operation.

Here we reconsider an old but largely overlooked idea originally proposed by David JC \citet{MacKay1998} in a different setting (arguably the inverse of the Deep Learning setting), which transforms a Dirichlet distribution into a Gaussian. When Dirichlet distributions are transformed with the inverse-softmax function, its shape effectively approximates a Gaussian. The inverse of this approximation, which will be called the \emph{Laplace Bridge} (LB) here \citep{KernelTopicModels2012}, can also in principle analytically map the parameters of a Gaussian distribution onto those of a Dirichlet distribution. Given a Gaussian distribution over the logits of a NN, one can thus efficiently obtain an approximate Dirichlet distribution over the softmax outputs. However, the bridge was previously used to map in the Gaussian to Dirichlet direction. The inverse direction of the vanilla LB has some limitations, arguably caused by the larger state-space of Gaussian relative to the Dirichlet exponential family. 

\begin{figure*}[!htb]
    %\vspace{-0.5em}
    \centering
    \includegraphics[width=\textwidth]{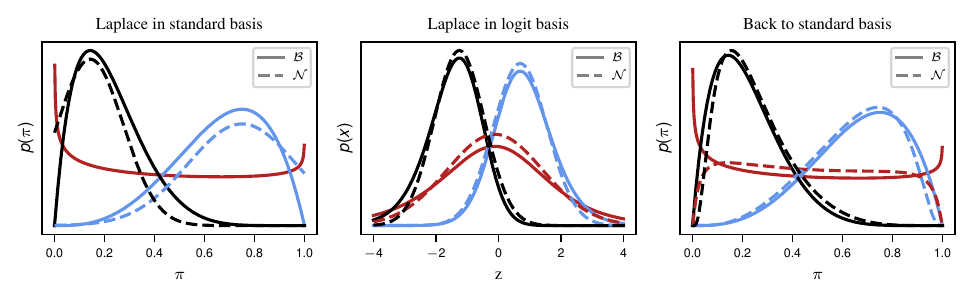}
    \vspace{-2.5em}
    \caption{(Adapted from \citet{KernelTopicModels2012}). Visualization of the Laplace Bridge for the Beta distribution (1D special case of the Dirichlet) for three sets of parameters. \textbf{Left:} ``Generic'' Laplace approximations of standard Beta distributions by Gaussians. Note that the Beta Distribution (red) does not have a valid approximation because its Hessian is not positive semi-definite. \textbf{Middle:} Laplace approximation to the same distributions after basis transformation through the softmax \eqref{eq:dirichlet_softmax}. The transformation makes the distributions ``more Gaussian'' (i.e.~uni-modal, bell-shaped, with support on the real line), thus making the Laplace approximation more accurate. \textbf{Right:} The same Beta distributions, with the back-transformation of the Laplace approximations from the middle figure to the simplex, yielding an improved approximate distribution. In contrast to the left-most image, the dashed lines now actually are probability densities (they integrate to $1$ on the simplex).}
    \label{fig:1D_Laplace_bridge}
    %\vspace{-0.75em}
\end{figure*}

Our contributions are a) We analyze these limits and suggest a solution that allows for the practical application of the LB. b) We show how the result can be used in the context of BNNs (see Figure \ref{fig:LB_for_BNN_overview} and \ref{fig:2D_LB_for_NNs}). 
c) We empirically evaluate the quality of this approximation, its speed-up, and its performance for out-of-data distribution tasks.
d) Finally, we show a use case on ImageNet, leveraging the analytic properties of Dirichlets to improve the popular top-$k$ metric through uncertainties. 
%%%%%%%%%%%%%%%%%%%%%%%%
\section{The Laplace Bridge}
\label{sec:background}
%%%%%%%%%%%%
Laplace approximations\footnote{For clarity: Laplace approximations are \emph{also} one out of several possible ways to construct a Gaussian approximation to the weight posterior of a NN, by constructing a second-order Taylor approximation of the empirical risk at the trained weights. This is \emph{not} the way they are used in this section. The LB is agnostic to how the input Gaussian distribution is constructed as it maps parameters. It could, e.g., also be constructed as a variational approximation, or the moments of Monte Carlo samples.}\citep{MacKay1992,laplace2021} are a popular and lightweight method to approximate general probability distributions $q(\vx)$ with a Gaussian $\N(\vx | \vmu, \mSigma)$ when $q(\vx)$ is twice differentiable and the Hessian at the mode is positive definite. They set $\vmu$ to a mode of $q$, and $\mSigma=-(\nabla^2 \log q(\vx) \vert_\vmu)^{-1}$, the inverse Hessian of $\log q$ at that mode. This scheme can work well if the true distribution is unimodal and defined on the real vector space.

The Dirichlet distribution, which has the density function
\begin{equation}\label{eq:dirichlet}
    \mathrm{Dir}(\vpi | \valpha) := \frac{\Gamma \left( \sum_{k=1}^K \alpha_k \right)}{\prod_{k=1}^K \Gamma(\alpha_k)} \prod_{k=1}^K \pi_k^{\alpha_k-1} \, ,
\end{equation}
is defined on the probability simplex and can be ``multimodal'' in the sense that the distribution diverges in the $k$-corner of the simplex when $\alpha_k < 1$\textbf{}. This precludes a Laplace approximation, at least in the na\"ive form described above. However, \citet{MacKay1998} noted that both can be fixed elegantly by a change of variable (Figure \ref{fig:1D_Laplace_bridge}). Details of the following argument can be found in Appendices \ref{appendix_B_LB} and \ref{appendix_C_inversion}. Consider the $K$-dimensional variable $\vpi \sim \mathrm{Dir}(\vpi | \valpha)$ defined as the softmax of $\vz\in\mathbb{R}^K$:
\begin{equation}
    \pi_k(\vz) := \frac{\exp(z_k)}{\sum_{l=1}^K \exp(z_l)} \, ,
\end{equation}
for all $k = 1, \dots, K$. We will call $\vz$ the logit of $\vpi$. When expressed as a function of $\vz$, the density of the Dirichlet in $\vpi$ has to be multiplied by the absolute value of the determinant of the Jacobian
\begin{equation}
    \det \frac{\partial \vpi}{\partial \vz}  = \prod_k \pi_k(z_k),
\end{equation}
thus removing the ``$-1$'' terms in the exponent: 
\begin{equation}\label{eq:dirichlet_softmax}
    \mathrm{Dir}_{\vz}(\vpi(\vz) | \valpha) := \frac{\Gamma \left( \sum_{k=1}^K \alpha_k \right)}{\prod_{k=1}^K \Gamma(\alpha_k)} \prod_{k=1}^K \pi_k(\vz)^{\alpha_k}
\end{equation}
This density of $\vz$, the Dirichlet distribution in the \emph{softmax basis}, can now be accurately approximated by a Gaussian through a Laplace approximation (see Figure \ref{fig:1D_Laplace_bridge}), yielding an analytic map from the parameter $\valpha\in\mathbb{R}_+ ^K$ to the parameters of the Gaussian ($\vmu \in\mathbb{R}^K$ and symmetric~positive~definite $\mSigma\in\mathbb{R}^{K\times K}$), given by
\begin{align}
\mu_k &= \log \alpha_k  - \frac{1}{K} \sum_{l=1}^{K} \log \alpha_l \, , \label{eq:mubridge}\\
\Sigma_{k\ell} &= \delta_{k\ell}\frac{1}{\alpha_k} - \frac{1}{K}\left[\frac{1}{\alpha_k} + \frac{1}{\alpha_\ell} - \frac{1}{K}\sum_{u=1} ^K \frac{1}{\alpha_u} \right]. \label{eq:Sigmabridge}
\end{align}
The corresponding derivations require care because the Gaussian parameter space is evidently larger than that of the Dirichlet and not fully identified by the transformation.
A pseudo-inverse of this map was provided as a side result in \citet{KernelTopicModels2012}. It maps the Gaussian parameters to those of the Dirichlet as
\begin{equation} \label{eq:alpha_transform}
    \alpha_k = \frac{1}{\Sigma_{kk}}\left(1 - \frac{2}{K} + \frac{e^{\mu_k}}{K^2}\sum_{l=1}^K e^{-\mu_l} \right) \,
\end{equation}
(this equation ignores off-diagonal elements of $\mSigma$, more discussion in Appendix \ref{appendix_C_inversion}). 
Together, Eqs.~\eqref{eq:mubridge}, \eqref{eq:Sigmabridge} and \eqref{eq:alpha_transform} will be called the \emph{Laplace Bridge}. For Bayesian Deep Learning, we only use Equation \eqref{eq:alpha_transform} which maps from $\boldsymbol{\mu}, \boldsymbol\Sigma$ to $\boldsymbol\alpha$. 
Even though the LB implies a reduction of the distribution's expressiveness, we show in \Cref{sec:method} that this map is still sufficiently accurate.

%%%%%%%%%%%%%%%%%%%%%%%%
\section{The Laplace Bridge for BNNs}
\label{sec:method}
The Laplace Bridge can be applied to any NN setup that maps from a Gaussian to probabilities by using the softmax. Throughout this paper, we use a last-layer Laplace approximation of the network as successfully used e.g. by \citet{ScalableBayesianOptimizationDNNs2015,kristiadi2020being}. It is given by
\begin{equation} \label{eq:logit_dist_last_layer}
    q(\vz \vert \vx) \approx \N(\vz | \bm{\mu}_{\mathbf{W}^{(L)}} \phi(\vx), \phi(\vx)^T \mSigma_{\mathbf{W}^{(L)}} \phi(\vx)) \, ,
\end{equation}
where $\phi(\vx)$ denotes the output of the first $L-1$ layers, $\boldsymbol{\mu}_{\mathbf{W}^{(l)}}$ is the maximum a posteriori (MAP) estimate for the weights of the last layer, and $\mSigma_{\mathbf{W}^{(l)}}$ is the inverse of the negative loss Hessian w.r.t. $\mathbf{W}^{(l)}$, $\mSigma_{\mathbf{W}^{(L)}} = -(\nabla^2_{\mathbf{W}^{(L)}} \L)^{-1}$ around the MAP estimate $\mathbf{W}^{(L)}$.
Even though last-layer Laplace approximations only use uncertainty from the last linear layer, they empirically perform as well as full Laplace approximations \citep{kristiadi2020being}. Furthermore, they allow for very fast inference, thus being a good match for the LB. We use diagonal and Kronecker approximations to the Hessian (see Appendix \ref{appendix_D_experiments}). 

Using the LB we can \emph{analytically} approximate the density of the softmax-Gaussian random variable that is the output of the BNN as a Dirichlet rather than using many samples. 
As shown in Eq.~\eqref{eq:alpha_transform}, it requires $\mathcal{O}(K)$ computations to construct the $K$ parameters $\alpha_k$ of the Dirichlet. In contrast, MC-integration has computational costs of $\mathcal{O}(MJ)$, where $M$ is the number of samples and $J$ is the cost of sampling from $q(\vz \vert \vx)$ (typically $J$ is of order $K^2$ after an initial $\mathcal{O}(K^3)$ operation for a matrix decomposition of the covariance). The MC approximation has the usual sampling error of $\mathcal{O}(1/\sqrt{M})$, while the LB has a fixed but small error (empirical comparison in \Cref{subsec:exp3_time}). This means that computing the LB is faster than drawing a single MC sample while yielding a full distribution.

Further benefits of this approximation arise from the convenient analytical properties of the Dirichlet exponential family. For example, a point estimate of the posterior predictive distribution is directly given by the Dirichlet's mean,
\begin{equation}\label{eq:dirichlet_mean}
    \mathbb{E}[\vpi] = \left( \frac{\alpha_1}{\sum_{l=1}^K \alpha_l}, \dots, \frac{\alpha_K}{\sum_{l=1}^K \alpha_l} \right)^\top \, .
\end{equation}
This removes the necessity for MC integration and can be computed analytically.
Additionally, Dirichlets have Dirichlet marginals: If $p(\vpi) = \mathrm{Dir}(\vpi | \valpha)$, then
\begin{equation}
      \label{eq:dirichlet_marginal}
     p \left( \pi_1,\dots,\pi_j,\sum_{k>j}\pi_k \right)\\ = \mathrm{Dir} \left( \alpha_1,\dots,\alpha_j,\sum_{k>j}\alpha_k \right) \, .
\end{equation}
Thus marginal distributions of arbitrary subsets of outputs (including binary marginals) can be computed in closed-form.

An additional benefit of the LB for BNNs is that it is more flexible than an MC-integral. If we let $p(\vpi)$ be the distribution over $\vpi := \mathrm{softmax}(\vz) := [e^{z_1}/\sum_{l} e^{z_l}, \dots, e^{z_K}/\sum_{l} e^{z_l}]^\top$, then the MC-integral can be seen as a ``point-estimate'' of this distribution since it approximates $\mathbb{E}[\vpi]$. In contrast, the Dirichlet distribution $\mathrm{Dir}(\vpi | \valpha)$ approximates the distribution $p(\vpi)$. Thus, the LB enables tasks that can be done only with a distribution but not a point estimate. For instance, one could ask ``what is the distribution of the softmax output of the first $L$ classes?'' when one is dealing with $K$-class ($L < K$) classification. Since the marginal distribution can be computed analytically with Eq. \eqref{eq:dirichlet_marginal}, the LB provides a convenient yet cheap way of answering this question.
%%%%%%%%%%%%%%%%%%%%%%%%
\section{Limitations of the Laplace Bridge}
\label{sec:lb_limits}
\begin{figure*}[t]
    %\vspace{-0.5em}
    \centering
    \includegraphics[width=\textwidth]{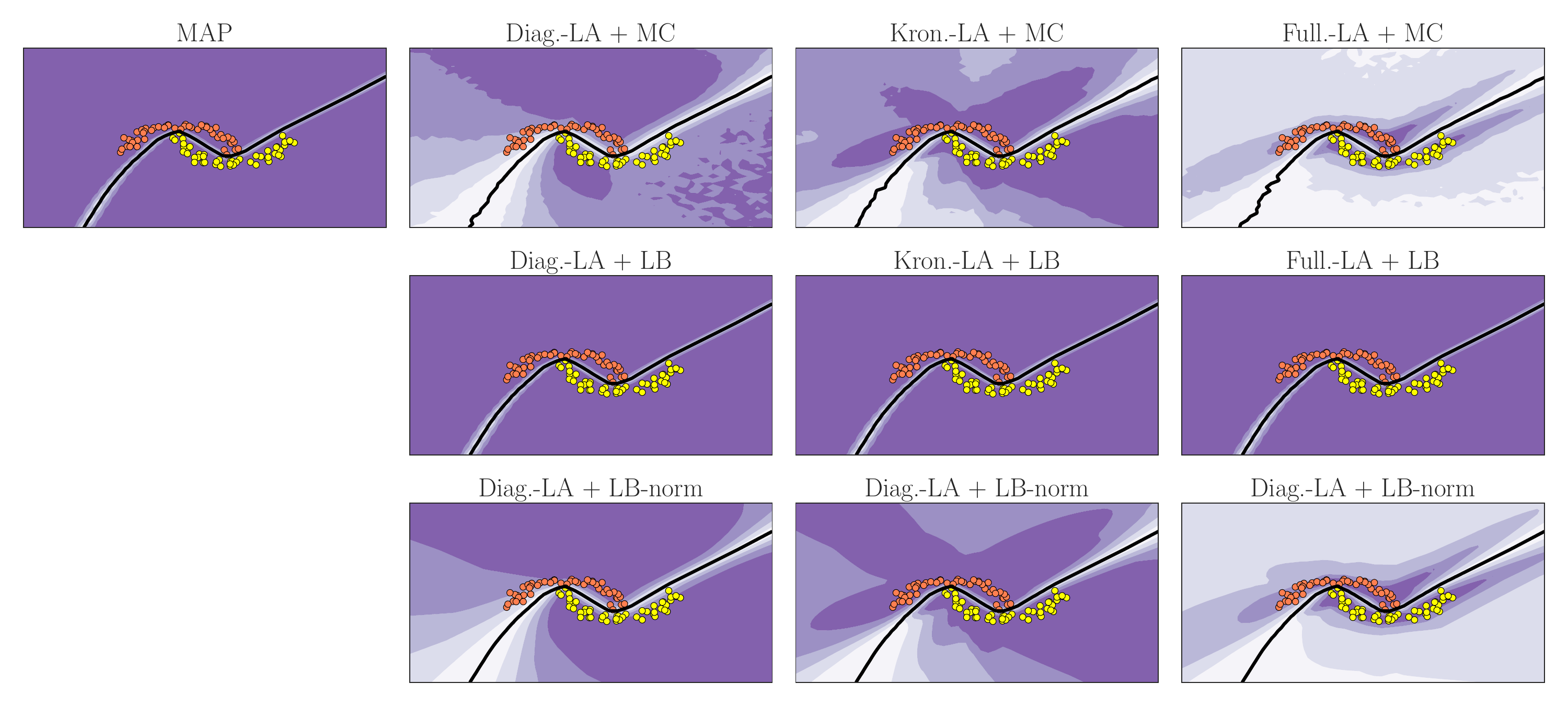}
    \vspace{-2em}
    \caption{\textbf{Left column:} vanilla MAP estimate which is overconfident. \textbf{Top row:} mean of softmax applied to Gaussian samples. \textbf{Middle row:} mean of the vanilla LB. \textbf{Bottom row:} mean of the corrected LB. The vanilla LB yields overconfident prediction far from the data. Our proposed correction fixes this issue, making the LB's approximation close to MC.}
    \label{fig:2D_LB_for_NNs}
    %\vspace{-0.75em}
\end{figure*}

\begin{figure}[!t]
    %\vspace{-0.75em}
    \centering
    \includegraphics{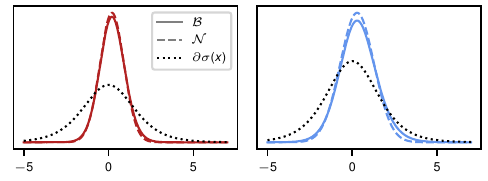}
    \includegraphics{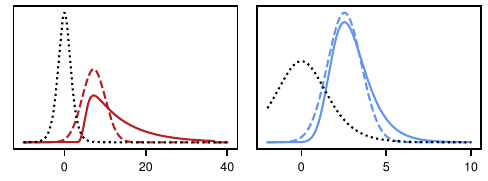}
    \vspace{-0.75em}
    \caption{
    %\textbf{Limitations of the LB:} 
    In most scenarios (upper row) the LB provides a good fit. However, in some high-variance scenarios (lower row) the softmax-Dirichlet becomes asymmetric and thus the Gaussian is a suboptimal fit. We propose a correction (right column) that projects the Gaussian into a lower-variance region before applying the LB. This can be understood as ``pulling back'' the Dirichlet to the dynamic of the logistic function (indicated here by its derivative $\partial \sigma$) and thus yields a better approximation.
    }
    \label{fig:2D_limitations}
    %\vspace{-0.75em}
\end{figure}

There are two limitations to applying the LB as presented in Equation \eqref{eq:alpha_transform}. 
First, the LB assumes that the random variable of the Gaussian sums to zero due to the difference in degrees of freedom between Dirichlet and Gaussian (see Appendix \ref{appendix_C_inversion}). Thus, we have to add a correction that projects from any arbitrary Gaussian to one that fulfills this constraint. The resulting Gaussian (see Appendix \ref{appendix_A}) is
\begin{equation} \label{eq:proj_gaussian}
    \mathcal{N}\left(\vx \vert \mu - \frac{\Sigma \mathbf{1} \mathbf{1}^\top \mu}{\mathbf{1}^\top \Sigma \mathbf{1}}, \Sigma - \frac{\Sigma \mathbf{1} \mathbf{1}^\top \Sigma}{\mathbf{1}^\top \Sigma \mathbf{1}} \right)
\end{equation}
where $\mathbf{1}$ is the one-vector of size $K$. 

Second, the softmax-Dirichlet distribution is asymmetric for extremely sparse cases (see Figure \ref{fig:2D_limitations}). 
These arise in regions where the logistic transform (the 1D special case of the softmax) is nearly flat (as indicated by its derivative in Figure \ref{fig:2D_limitations}). Therefore, the LA is suboptimal in these high-variance cases. 

This limitation can also be explained by looking at Equation \eqref{eq:alpha_transform}.
We observe that $\Sigma$ contributes linearly to $\alpha$ with $\frac{1}{\Sigma_{kk}}$ while $\mu$ contributes exponentially with $\exp(\mu_k)$. For settings where $\Sigma$ is small, this doesn't have a large effect. However, when $\Sigma_{kk}$ and $\mu_k$ grow the LB results differ from softmax Gaussian samples. In the LB, the resulting $\alpha$ is dominated by the mean and the linear influence of the variance cannot correct sufficiently. For Monte Carlo sampling, on the other hand, the result is mostly determined by the large variance and then amplified through the softmax. Our proposed normalization to the LB reduces this effect (see Figure \ref{fig:2D_LB_for_NNs}).

In BNNs, we often encounter such cases, especially far away from the data (see Figure \ref{fig:2D_LB_for_NNs} top). Therefore, we propose an additional correction for practical purposes: 
\begin{align}
    c &= v_\text{mean}(\Sigma) \cdot \frac{1}{\sqrt{K/2}} \\
    \mu' &= \frac{\mu}{\sqrt{c}} \\
    \Sigma' &= \frac{\Sigma}{c}
\end{align}
where $v_\text{mean}(\Sigma)$ denotes the mean variance of $\Sigma$, $v_{\text{mean}}(\Sigma) = \sum_i \Sigma_{ii}$. The factor of $\frac{1}{\sqrt{K/2}}$ is added because we found that higher dimensionalities require less correction. Since our correction is just a rescaling, the zero-sum constrained is still fulfilled.
This normalization that can be understood as ``pulling back'' the distribution into a space where it is symmetric has higher approximation quality. This correction is applied after the zero-sum constraint correction. 

We want to point out that our correction is motivated by experimentation and the theoretical insights detailed above. There is no theoretical derivation from first principles for the correction. We provide additional explanations and figures in Appendix \ref{appendix_A}. 

Throughout the paper, we will call this normalizing correction \textit{LB-norm} and explicitly state when we use it. Otherwise, we will use the vanilla version with zero-sum correction. 
%%%%%%%%%%%%%%%%%%%%%%%%
\section{Related Work}
\label{sec:related}
%%%%%%%%%%%%%%%%%%%%%
\begin{table*}[ht!]
    \scriptsize
    \fontsize{8}{8}\selectfont
    \setlength{\tabcolsep}{1.5pt}
    \centering
    \caption{OOD detection results. In all scenarios, the Laplace Bridge (LB) or its normalized version yield comparable results to the MC estimate while being much faster. 
    For MC experiments, we draw 100 samples. 
    %Results are averages from five runs with different seeds.
    }
    \vspace{-0.75em}
    %#######################################
    %\resizebox{\textwidth}{!}{% use resizebox with textwidth
    \begin{tabular}{l  l || c c c c c c | c  c  c  c  c c}
         \toprule
         & & \multicolumn{2}{c}{\textbf{Diag.-LA + MC}} & \multicolumn{2}{c}{\textbf{Diag.-LA + LB}} & \multicolumn{2}{c}{\textbf{Diag.-LA + LB-norm}} &\multicolumn{2}{c}{\textbf{Kron.-LA + MC}} &  \multicolumn{2}{c}{\textbf{Kron.-LA + LB}} & \multicolumn{2}{c}{\textbf{Kron.-LA + LB-norm}} \\
         \textbf{Train} & \textbf{Test} & \textbf{ECE} $\downarrow$ & \textbf{AUROC} $\uparrow$& \textbf{ECE} $\downarrow$ & \textbf{AUROC} $\uparrow$& \textbf{ECE} $\downarrow$& \textbf{AUROC} $\uparrow$& \textbf{ECE}$\downarrow$ & \textbf{AUROC} $\uparrow$ & \textbf{ECE} $\downarrow$ & \textbf{AUROC} $\uparrow$& \textbf{ECE} $\downarrow$ & \textbf{AUROC} $\uparrow$ \\
         \midrule
        MNIST &   FMNIST &             \textbf{0.464} &               0.975 &        0.478 &          \textbf{0.981} &             0.498 &               0.951 &             0.390 &               0.987 &        0.553 &          0.977 &             \textbf{0.364} &               \textbf{0.990} \\
   MNIST & notMNIST &             0.396 &               \textbf{0.965} &        0.600 &          0.930 &             \textbf{0.360} &               0.955 &             0.366 &               0.974 &        0.634 &          0.912 &             \textbf{0.294} &               \textbf{0.986} \\
   MNIST &   KMNIST &             0.429 &               \textbf{0.974} &        0.617 &          0.949 &             \textbf{0.391} &               0.970 &             0.374 &               0.985 &        0.619 &          0.956 &             \textbf{0.328} &               \textbf{0.991} \\
         \midrule
          CIFAR10 & CIFAR100 &             0.379 &               \textbf{0.887} &        0.691 &          0.859 &             \textbf{0.220} &               0.883 &             0.577 &               \textbf{0.878} &        0.670 &          0.855 &             \textbf{0.558} &               0.866 \\
 CIFAR10 &     SVHN &             0.309 &               \textbf{0.948} &        0.652 &          0.928 &             \textbf{0.155} &               \textbf{0.948} &             0.447 &               0.955 &        0.635 &          0.924 &             \textbf{0.327} &               \textbf{0.965} \\
         \midrule
        SVHN & CIFAR100 &             \textbf{0.615} &               0.957 &        0.667 &          \textbf{0.962 }&             0.679 &               0.944 &             0.583 &               \textbf{0.959} &        0.659 &          0.962 &             \textbf{0.575} &               0.953 \\
    SVHN &  CIFAR10 &             \textbf{0.600} &               0.958 &        0.659 &          \textbf{0.960} &             0.662 &               0.947 &             0.567 &               \textbf{0.960} &        0.651 &          0.959 &             \textbf{0.556} &               0.955 \\
         \midrule
        CIFAR100 &  CIFAR10 &             0.474 &               0.788 &        \textbf{0.239} &          \textbf{0.791} &             0.834 &               0.757 &             0.479 &               0.787 &        \textbf{0.202} &          \textbf{0.790} &             0.855 &               0.749 \\
CIFAR100 &     SVHN &             0.470 &               0.795 &        \textbf{0.207} &          \textbf{0.815} &             0.842 &               0.748 &             0.469 &               0.798 &        \textbf{0.183} &          \textbf{0.807} &             0.849 &               0.761 \\
         \bottomrule
    \end{tabular}
    %}
    \label{tab:experiments_table}
    %\vspace{-0.75em}
\end{table*}
%%%%%%%%%%%%%
In BNNs, analytic approximations of posterior predictive distributions have attracted a great deal of research. In the binary classification case, for example, the probit approximation \citep{PhDGibbs, LuProbit} has been proposed already in the 1990s \citep{spiegelhalter1990sequential,mackay1992evidence}. However, while there exist some bounds \citep{michalis2016one} and approximations of the expected log-sum-exponent function \citep{ahmed2007tight,braun2010variational}, in the multi-class case, obtaining a good analytic approximation of the expected softmax function under a Gaussian measure is an open problem. Our LB can be used to produce a close analytical approximation of this integral. 
It thus furthers the trend of sampling-free solutions within Bayesian Deep Learning \citep[etc.]{DeterministicVI2018, haussmann2019BEDLwithPAC}. The crucial difference is that, unlike these methods, the LB approximates the full distribution over the softmax outputs of a deep network.

Previous approaches proposed to model the distribution of softmax outputs of a network directly. Similar to the LB, \citet{NIPS2018PriorNetworks,NIPS2019PriorNetworks_improved,NIPS2018EvidentialDL} proposed to use the Dirichlet distribution to model the posterior predictive for non-Bayesian networks. They further proposed novel training techniques in order to directly learn the Dirichlet. Additionally, different work on Distillation \citep{malinin2019ensemble, vadera2020generalized} takes larger models and distills them into a smaller one. The result of some distillation methods is a Dirichlet similar to the LB. We compare against prior nets in the experiments. 

In contrast, the LB tackles the problem of approximating the distribution over the softmax outputs of the ubiquitous Gaussian-approximated BNNs \citep[etc]{Graves2011VB,Blundell2015WeightUI,louizos_structured_2016,sun_learning_2017} without any additional training procedure. Therefore the LB can, for example, be used with pre-trained weights on large datasets while prior networks and distillation usually require training from scratch.

%%%%%%%%%%%%%%%%%%%%%%%%
\section{Experiments}
\label{sec:experiments}
%%%%%%%%%%%%%%%%%%%%%
\begin{table*}[htb!]
    \scriptsize
    \fontsize{9}{10}\selectfont
    \setlength{\tabcolsep}{4pt}
    \centering
    \caption{Comparison of the extended probit approximation with the normalized version of the LB norm. While the probit approximation performs well on in-dist problems, the LB norm is better on out-of-distribution tasks.
    }
    \vspace{-0.75em}
    %#######################################
    %\resizebox{\textwidth}{!}{% use resizebox with textwidth
    \begin{tabular}{l  l || c c c c c | c c  c  c  c }
         \toprule
         & & \multicolumn{5}{c}{\textbf{Diag Probit}} & \multicolumn{5}{c}{\textbf{Diag LB norm}} \\
         \textbf{Train} & \textbf{Test} & \textbf{MMC} $\downarrow$ & \textbf{AUROC} $\uparrow$& \textbf{NLL} $\downarrow$ & \textbf{ECE} $\downarrow$ & \textbf{Brier} $\downarrow$ & \textbf{MMC} $\downarrow$ & \textbf{AUROC} $\uparrow$& \textbf{NLL} $\downarrow$ & \textbf{ECE} $\downarrow$ & \textbf{Brier} $\downarrow$\\
         \midrule
  MNIST &    MNIST &            \textbf{0.967} &              - &            0.050 &            0.024 &              0.002 &             0.944 &               - &             0.078 &             0.045 &               0.003 \\
  MNIST &   FMNIST &            0.597 &              \textbf{0.971} &            3.827 &            0.523 &              0.128 &             \textbf{0.589} &               0.951 &             \textbf{3.538} &             \textbf{0.498} &               \textbf{0.124} \\
  MNIST & notMNIST &            0.616 &              \textbf{0.958} &            3.839 &            0.488 &              0.123 &             \textbf{0.492} &               0.955 &             \textbf{3.070} &             \textbf{0.360} &               \textbf{0.111} \\
  MNIST &   KMNIST &            0.580 &              0.969 &            4.276 &            0.489 &              0.126 &             \textbf{0.484} &               \textbf{0.970} &             \textbf{3.288} &             0.391 &               \textbf{0.115} \\
  \midrule
CIFAR10 &  CIFAR10 &            \textbf{0.869} &              - &            0.237 &            0.083 &              0.009 &             0.517 &               - &             0.727 &             0.433 &               0.029 \\
CIFAR10 & CIFAR100 &            0.589 &              0.882 &            3.334 &            0.485 &              0.123 &             \textbf{0.319} &               \textbf{0.883} &             \textbf{2.590} &             \textbf{0.220} &               \textbf{0.099} \\
CIFAR10 &     SVHN &            0.510 &              0.946 &            3.097 &            0.394 &              0.114 &             \textbf{0.273} &               \textbf{0.948} &             \textbf{2.457} &             \textbf{0.155} &               \textbf{0.094} \\
\bottomrule
    \end{tabular}
    %}
    \label{tab:probit_table_diag}
    \vspace{-1em}
\end{table*}
%%%%%%%%%%%%%%%%%%%%
\begin{table*}[ht!]
    \scriptsize
    \fontsize{8}{8}\selectfont
    \setlength{\tabcolsep}{1.5pt}
    \centering
    \caption{Comparison of last-layer vs. full-layer Laplace approximation. Last-layer results are in the upper half and full-layer results are in the bottom half. We find that, as expected, full-layer results are slightly better than for the last-layer approximation.
    %Results are averages from five runs with different seeds.
    }
    \vspace{-0.75em}
    %#######################################
    %\resizebox{\textwidth}{!}{% use resizebox with textwidth
    \begin{tabular}{l  l || c c c c c c | c  c  c  c  c c}
         \toprule
         & & \multicolumn{2}{c}{\textbf{Diag.-LA + MC}} & \multicolumn{2}{c}{\textbf{Diag.-LA + LB}} & \multicolumn{2}{c}{\textbf{Diag.-LA + LB-norm}} &\multicolumn{2}{c}{\textbf{Kron.-LA + MC}} &  \multicolumn{2}{c}{\textbf{Kron.-LA + LB}} & \multicolumn{2}{c}{\textbf{Kron.-LA + LB-norm}} \\
         \textbf{Train} & \textbf{Test} & \textbf{ECE} $\downarrow$ & \textbf{AUROC} $\uparrow$& \textbf{ECE} $\downarrow$ & \textbf{AUROC} $\uparrow$& \textbf{ECE} $\downarrow$& \textbf{AUROC} $\uparrow$& \textbf{ECE}$\downarrow$ & \textbf{AUROC} $\uparrow$ & \textbf{ECE} $\downarrow$ & \textbf{AUROC} $\uparrow$& \textbf{ECE} $\downarrow$ & \textbf{AUROC} $\uparrow$ \\
         \midrule
        \midrule
MNIST &   FMNIST &             0.464 &               0.975 &        0.478 &          0.981 &             0.498 &               0.951 &             0.390 &               0.987 &        0.553 &          0.977 &             0.364 &               0.990 \\
MNIST & notMNIST &             0.396 &               0.965 &        0.600 &          0.930 &             0.360 &               0.955 &             0.366 &               0.974 &        0.634 &          0.912 &             0.294 &               0.986 \\
MNIST &   KMNIST &             0.429 &               0.974 &        0.617 &          0.949 &             0.391 &               0.970 &             0.374 &               0.985 &        0.619 &          0.956 &             0.328 &               0.991 \\
\midrule
MNIST &   FMNIST &             0.317 &               0.980 &        0.322 &          0.990 &             0.123 &               0.986 &             0.288 &               0.985 &        0.528 &          0.980 &             0.135 &               0.991 \\
MNIST & notMNIST &             0.280 &               0.960 &        0.566 &          0.924 &             0.126 &               0.952 &             0.282 &               0.958 &        0.629 &          0.915 &             0.171 &               0.973 \\
MNIST &   KMNIST &             0.309 &               0.976 &        0.557 &          0.955 &             0.112 &               0.972 &             0.279 &               0.981 &        0.615 &          0.958 &             0.152 &               0.986 \\
\bottomrule
    \end{tabular}
    %}
    \label{tab:experiments_last_vs_full}
    %\vspace{-0.75em}
\end{table*}
%%%%%%%%%%%%%
%%%%%%%%%%%%%%%%%%%%%
\begin{table*}[htb!]
    \scriptsize
    \fontsize{9}{10}\selectfont
    \setlength{\tabcolsep}{4pt}
    \centering
    \caption{Comparison of Prior Networks with the normalized version of the LB norm. PNs consistently outperform the LB. For discussion see main text.
    }
    \vspace{-0.75em}
    %#######################################
    %\resizebox{\textwidth}{!}{% use resizebox with textwidth
    \begin{tabular}{l  l || c c c c c | c c  c  c  c }
         \toprule
         & & \multicolumn{5}{c}{\textbf{Prior Network}} & \multicolumn{5}{c}{\textbf{Diag LB norm}} \\
         \textbf{Train} & \textbf{Test} & \textbf{MMC} $\downarrow$ & \textbf{AUROC} $\uparrow$& \textbf{ECE} $\downarrow$ & \textbf{NLL} $\downarrow$ & \textbf{Brier} $\downarrow$ & \textbf{MMC} $\downarrow$ & \textbf{AUROC} $\uparrow$& \textbf{ECE} $\downarrow$ & \textbf{NLL} $\downarrow$ & \textbf{Brier} $\downarrow$\\
         \midrule
  MNIST &    MNIST &            0.802 &  - & 0.184 & 0.246 &  0.008  &             \textbf{0.944} &               - &           \textbf{0.045} &           \textbf{0.078} &                   \textbf{0.003} \\
  MNIST &   FMNIST &            \textbf{0.273} &  \textbf{0.995} & \textbf{0.212} & \textbf{2.659} &  \textbf{0.098}  &             0.589 &               0.951 &                0.498 &       3.538 &                  0.124 \\
  MNIST & notMNIST &      \textbf{0.447} &  0.938 & \textbf{0.314} & \textbf{2.962} &  \textbf{0.105} &             0.492 &               \textbf{0.955} &                  0.360 &        3.070 &               0.111 \\
  MNIST &   KMNIST &          \textbf{0.372} &  \textbf{0.976} & \textbf{0.261} & \textbf{3.142} &  \textbf{0.104} &             0.484 &               0.970 &                 0.391 &         3.288 &               0.115 \\
\bottomrule
    \end{tabular}
    %}
    \label{tab:PN_vs_diag_LB_norm}
    \vspace{-1em}
\end{table*}
%%%%%%%%%%%%%%%%%%%%%%%%%
We conduct multiple experiments.
Firstly, we compare the LB to the MC-integral on a 2D toy example (\Cref{subsec:exp1_2D_toy}). 
Secondly, we apply the same comparison to out-of-distribution (OOD) detection in many settings (\Cref{subsec:exp2_numbers}).
Thirdly, we compare the commonly used probit approximation to the LB in \cref{subsec:exp_probit_lb}
Fourthly, we compare their computational cost and contextualize the speed-up for the prediction process in \Cref{subsec:exp3_time}.
Finally, in \Cref{subsec:exp4_imagenet}, we present analysis on ImageNet \citep{ImageNet2015} to demonstrate the scalability of the LB and the advantage of having a full Dirichlet distribution over softmax outputs.
We extended Laplace torch \citep{laplace2021} for the experiments. Code can be found in the accompanying GitHub repository.\footnote{\url{https://github.com/mariushobbhahn/LB_for_BNNs_official}} 

For all experiments, a last-layer Laplace approximation has been applied. This scheme has been successfully used by \citet{ScalableBayesianOptimizationDNNs2015,kristiadi2020being}. We use diagonal and Kronecker-factorized (KFAC)\citep{ritter2018a, martens2015optimizing} approximations of the Hessian, since inverting the exact Hessian is too costly. A detailed mathematical explanation and setup of the experiments can be found in \cref{appendix_D_experiments}.
%justification vs VI and other methods
While the LB could also be applied to different approximations of a Gaussian posterior predictive such as Variational Inference \citep{Graves2011VB,Blundell2015WeightUI}, we used a Laplace approximation in our experiments to construct such an approximation. This is for two reasons: (i) it is one of the fastest ways to get a Gaussian posterior predictive and (ii) it can be applied to pre-trained networks which is especially useful for large problems such as ImageNet.
Nevertheless, we want to emphasize again that the LB can be applied to any Gaussian over the outputs independent of the way it was generated. 
%%%%%%%%%%%%%%%%%%
\setlength{\figwidth}{0.45\textwidth}
\setlength{\figheight}{0.17\textheight}
%%%%%%%%%%%%%%%%%%
\begin{figure*}[htb!]
    \centering
    %\vspace{-1em}
    \scriptsize
    \subfloat{
        \subfloat{\includegraphics[width=\figwidth,height=\figheight]{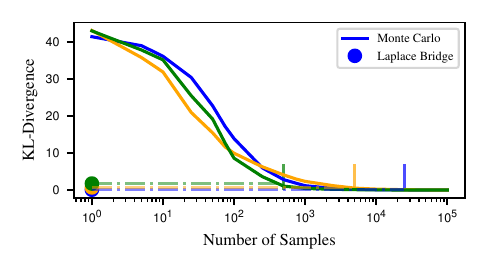}}
        \hspace{1em}
        \subfloat{\includegraphics[width=\figwidth,height=\figheight]{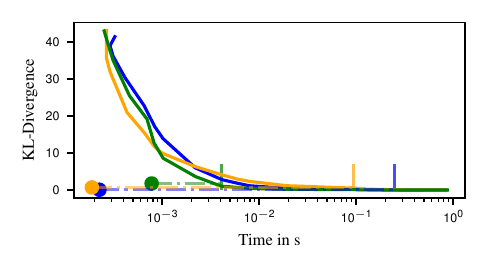}}
    }
    \vspace{-1em}
    \centering
    \caption{KL-divergence plotted against the number of samples (left) and wall-clock time (right). The Monte Carlo density estimation becomes as good as the LB after around $750$ to $10$k samples and takes at least $100$ times longer. The three lines (blue, yellow, green) represent three different sets of parameters. The short vertical bars indicate where the KL divergence of the samples overtake that of the LB.}
    \label{fig:KL_div_samples}
    \vspace{-1em}
\end{figure*}
%%%%%%%%%%%%%%%%%%%%%%%%%%%%%%%%%%%%%%%%
%\vspace{-0.5em}
\subsection{2D Toy example}
\label{subsec:exp1_2D_toy}
%%%%%%%%%%%%%%%%%%%%%%%%%%%%%%%%%%%%%%%%
We train a simple ReLU network on the 2D half-moon problems from scikit-learn \citep{scikit-learn}. As can be seen in Figure \ref{fig:2D_LB_for_NNs} the MAP estimate and vanilla LB are overconfident for the reasons discussed in \ref{sec:lb_limits} but the normalized version yields a near-perfect fit. 
%%%%%%%%%%%%%%%%%%%%%%%%%%%%%%%%%%%%%%%%
%\vspace{-0.5em}
\subsection{OOD detection}
\label{subsec:exp2_numbers}
%%%%%%%%%%%%%%%%%%%%%%%%%%%%%%%%%%%%%%%%
We compare the performance of the LB to the MC-integral (Diagonal and KFAC) on a standard OOD detection benchmark suite, to test whether the LB gives similar results to the MC sampling methods. Following prior literature, we use the standard expected calibration error (ECE) and area under the ROC-curve (AUROC) metrics \citep{HendycksOODBaseline}. 

For the exact setup, we refer the reader to Appendix \ref{appendix_D_experiments}.
We use the mean of the Dirichlet to obtain a comparable approximation to the MC-integral. The results are presented in Table \ref{tab:experiments_table}. 

We find that the results of the LB or its normalized version are comparable throughout the entire benchmark suite. Since the LB is much faster it can be a good replacement for MC in time-sensitive applications.

Furthermore, we compare the LB to prior networks (PNs) in Table \ref{tab:PN_vs_diag_LB_norm} since PNs also yield a Dirichlet distribution as an output on classification tasks. We find that PNs outperform the LB in most cases. However, we don't think this is a major problem since they have different aims and use cases. The LB creates a Dirichlet distribution on top of an already existing Gaussian model while PNs describe a training procedure and have to be trained from scratch. Thus, the primary comparison for the LB should be against sampling and other integral approximations like in Table \ref{tab:probit_table_diag}. 

Lastly, we compare the LB for a full-layer vs. last-layer Laplace approximation of the network in Table \ref{tab:experiments_last_vs_full}. We find that, as expected, the full-layer setting yield slightly better results. However, since the primary advantage of the LB is its speed, we think the natural fit for it is a last-layer approximation.
%%%%%%%%%%%%%%%%%%%%%%%%%%%%%%%%%%%%
\begin{table*}[!t]
    \centering
    \caption{Contextualization of the timings for the entire predictive process of a ResNet-18 on CIFAR-10. We see that with 1000 samples the forward pass only uses 6\% of the time whereas the sampling uses 94\%. In contrast the split for the LB is 96\% and 4\% respectively. We conclude that the LB provides a significant speed-up of the process as a whole.}
    \vspace{-0.75em}
    \resizebox{\textwidth}{!}{% use resizebox with textwidth
    \begin{tabular}{r | c | ccc c}
        \toprule
        \# samples in brackets  & Forward pass & $+$MC(1000) &  $+$MC(100) &  $+$MC(10) &   $+$Laplace Bridge \\
        \midrule
        Time in seconds & 0.300 $\pm$ 0.003 & 4.712 $\pm$ 0.063 & 0.488 $\pm$ 0.009 & 0.059 $\pm$ 0.001 & 0.013 $\pm$ 0.000 \\
        Fraction of overall time & 0.06/0.38/0.83/0.96 & 0.94 & 0.62 & 0.17 & 0.04 \\
        \bottomrule
    \end{tabular}
    }%end resizebox
    \vspace{-0.5em}
    \label{tab:timings_forward_vs_LB}
\end{table*}
%%%%%%%%%%%%%%%%%%%%%%%%%%%%%%%%%%%%%%%%
\setlength{\figwidth}{1\textwidth}
\setlength{\figheight}{0.16\textheight}

\begin{figure*}[t]
    %\vspace{-0.75em}
    \centering
    \includegraphics[width=\figwidth,height=\figheight]{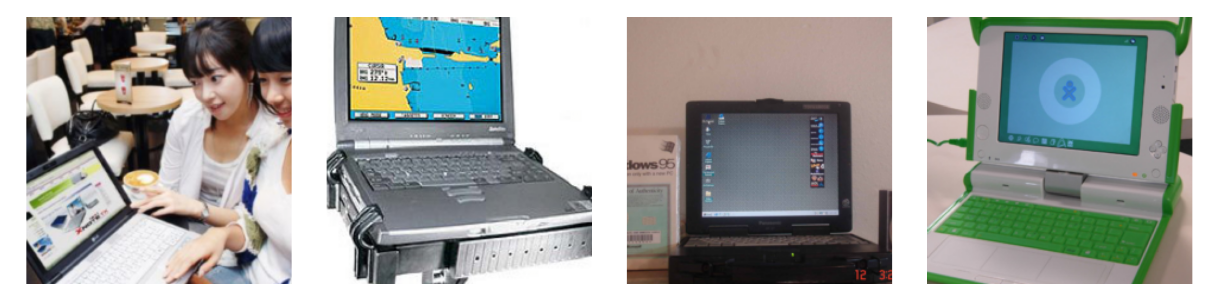}
    \includegraphics[width=\figwidth,height=\figheight]{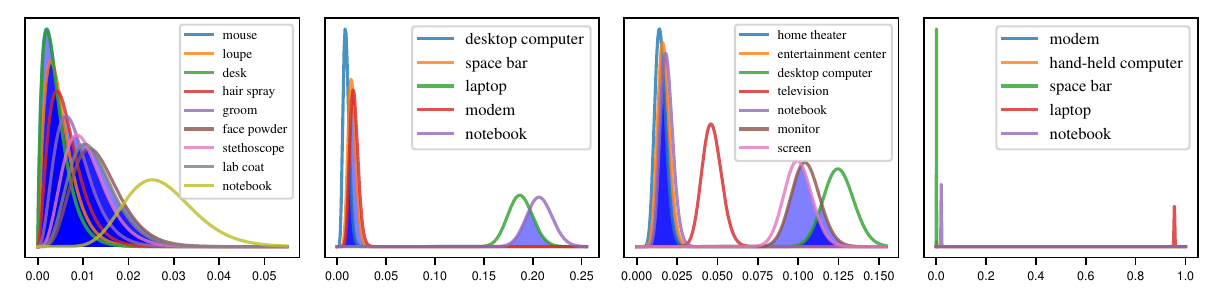}
    \caption{\textbf{Upper row:} images from the ``laptop'' class of ImageNet. \textbf{Bottom row:} Beta marginals of the top-$k$ predictions for the respective image. In the first column, the overlap between the marginal of all classes is large, signifying high uncertainty, i.e. the prediction is ``do not know''. In the second column, ``notebook'' and ``laptop'' have confident, yet overlapping marginal densities and therefore yield a top-$2$ prediction: ``either notebook or laptop''. In the third column ``desktop computer'', ``screen'' and ``monitor'' have overlapping marginal densities, yielding a top-$3$ estimate. The last case shows a top-$1$ estimate: the network is confident that ``laptop'' is the only correct label.
    }
    \label{fig:imagenet_betas}
    %\vspace{-0.75em}
\end{figure*}
%%%%%%%%%%%%%%%%%%%%%%%%%%%%%%%%%%%%%%%%%%%%%
\subsection{Comparison to the probit approximation}
\label{subsec:exp_probit_lb}
%%%%%%%%%%%%%%%%%%%%%
The multi-class probit approximation \citep{PhDGibbs, LuProbit} is a commonly used approximation for the softmax-Gaussian integral. We compare it to the diagonal normalized LB in Table \ref{tab:probit_table_diag}. We find that the LB norm outperforms the probit approximation in most OOD tasks. When we use a KFAC approximation of the Hessian, this trend still holds (see Table \ref{tab:probit_table_kfac} in Appendix \ref{appendix_D_experiments}).
\subsection{Time comparison}
\label{subsec:exp3_time}
%%%%%%%%%%%%%%%%%%%%%%%%%%%%%%%%%%%%%
We compare the computational cost of the density-estimated $p_\text{sample}$ distribution via sampling and the Dirichlet obtained from the LB $p_\text{LB}$ for approximating the true $p_\text{true}$ over MC-sampling. Different numbers of samples are drawn from the Gaussian, the softmax is applied and the KL-divergence between the histogram of the samples with the true distribution is computed. We use KL-divergences $D_\text{KL}(p_\text{true} \Vert p_\text{sample})$ and $D_\text{KL}(p_\text{true} \Vert p_\text{LB})$, respectively, to measure similarity between approximations and ground truth while the number of samples for $p_\text{sample}$ is increased exponentially. The true distribution $p_\text{true}$ is constructed via MC with $100$k samples. The experiment is conducted for three different Gaussian distributions over $\R^3$. Since the softmax applied to a Gaussian does not have an analytic form, the algebraic calculation of the approximation error is not possible and an empirical evaluation via sampling is the best option. The fact that there is no analytic solution is part of the justification for using the LB in the first place. 
%The logistic-Normal distribution \citep{logitNormal1980} is a different distribution and not useful for our purposes as it has no analytic expected value and different support.

\Cref{fig:KL_div_samples} suggests that the number of samples required such that the distribution $p_\text{sample}$ approximates the true distribution $p_\text{true}$ as good as the Dirichlet distribution obtained via the LB is large, i.e. somewhere between $750$ and $10$k. This translates to a wall-clock time advantage of at least a factor of $100$ before sampling becomes competitive in quality with the LB.

%LB is light-weight, i.e. easy and fast to apply.
To further demonstrate the low compute cost of the LB, we timed different parts of the process for our setup. On our hardware and setup, training a ResNet-18 on CIFAR10 over 130 epochs takes 71 minutes and 30 seconds. Computing a Hessian for the network from the training data can be done with {\sc BackPACK} \citep{dangel2020backpack} at the cost of one backward pass over the training data or around 29 seconds. This one additional backward pass is the only change to the training procedure compared to conventional training.
%LB actually makes a difference compared to the forward pass
Since the LB only applies to the last step of the prediction pipeline, it is important to compare it to a forward pass through the rest of the network. Re-using the ResNet-18 and CIFAR10 setup we measure the time in seconds for a forward pass, for the application of the LB, and for the sampling procedure with 10, 100, and 1000 samples. The resulting sum total time for the entire test set is given in Table \ref{tab:timings_forward_vs_LB}. We find that sampling takes up between 94\% (for 1000 samples) and 17\% (for 10 samples) of the entire prediction while the LB is only 4\%. Thus, the acceleration through the LB is a significant improvement for the prediction process as a whole, not only for a part of the pipeline.
%%%%%%%%%%%%%%%%%%%%%%%%%%%%%%%%%%%%%%%%%%%%%%%%%%%%%%%%%
\subsection{Uncertainty-aware output ranking on ImageNet}
\label{subsec:exp4_imagenet}

Due to the cost of sampling-based inference, classification on large datasets with many classes, like ImageNet, is rarely done in a Bayesian fashion. Instead, models for such tasks are often compared along a top-$k$ metric (e.g.~$k=5$).
%, i.e.~it is tested whether the correct class is within the five most probable estimates of the network. 

Although widely accepted, this metric has some pathologies: 
Depending on how close the point predictions are \emph{relative to their uncertainty},  the total number of likely class labels should be allowed to vary from case to case. Figure  \ref{fig:imagenet_betas} shows examples: In some cases (panel 2) the classifier is quite confident that the image in question belongs to one out of only two classes and all others are highly unlikely. In others (e.g.~panel 1), a larger set of hypotheses are all nearly equally probable.

The Laplace Bridge, in conjunction with the last-layer Laplace approximations, can be used to address this issue. To this end, the analytic properties of its Dirichlet prediction are particularly useful: Recall that
the marginal distribution $p(\pi_i,\sum_{j\neq i} \pi_j)$ over each component of a Dirichlet relative to all other components is  $\mathrm{Beta}(\alpha_i, \sum_{j\neq i} \alpha_j)$. 

We leverage this property to propose a simple \emph{uncertainty-aware top-$k$} decision rule inspired by statistical tests. Instead of keeping $k$ fixed, it uses the model's confidence to adapt $k$ (pseudo-code in Algorithm \ref{alg:ua-top-k}).

We begin by sorting the class predictions in order of their expected probability $\alpha_i$. Then we compute the Beta marginal of the most likely class. Now, we compute the overlap of the next marginal and add that class to the list iff the overlap is more than some threshold (e.g.~0.05). Continuing in this fashion, the algorithm terminates with a finite value $k\leq K$ of ``non-separated'' top classes. 
% This is inspired by hypotheses testing,  \citep[e.g.][]{nickerson2000null} or its Bayesian alternatives \citep{BayesianAltTTest2011}. 

The intuition behind this rule is that, if any Beta density overlaps with the most likely one more than the threshold of, say, $5\%$, the classifier cannot confidently predict one class over the other. Thus, all classes sufficiently overlapping with the top contender should be returned as the top estimates. 
%%%%%%%%%%%%%%%%%%%%
\begin{figure}[!t]
    \vspace{-0.75em}
    \centering
    \includegraphics[width=0.49\textwidth]{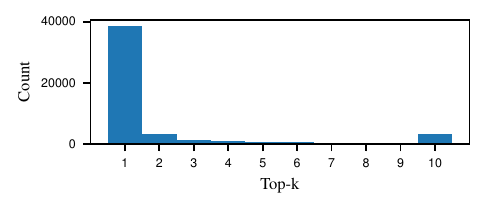}
    \vspace{-2em}
    \includegraphics[width=0.49\textwidth]{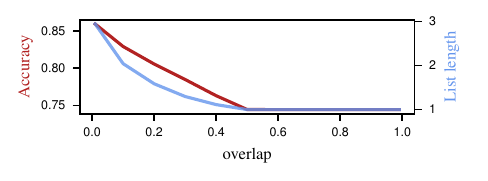}
    %\vspace{-0.75em}
    \caption{A histogram of ImageNet predictions' length using the proposed uncertainty-aware top-$k$. Results with more than 10 proposed classes have been put into the 10-bin for visibility.}
    \label{fig:imagenet_counts}
    %\vspace{-1em}
\end{figure}
%%%%%%%%%%%%%%%%%%

We evaluate this decision rule on the test set of ImageNet. The overlap is calculated through the inverse CDF\footnote{Also known as the quantile function or percent point function} of the respective Beta marginals. The original top-$1$ accuracy of DenseNet on ImageNet is $0.744$. In contrast, the uncertainty-aware top-k method yields accuracies of over $0.85$ while average list lengths stay below $3$ (see Figure \ref{fig:imagenet_counts}). Furthermore, we find that most of the predictions given by the uncertainty-aware metric still yielded a top-$1$ prediction.
This means that using uncertainty does not imply adding meaningless classes to the prediction.
Furthermore, there are non-negligibly many cases where $k$ equals to $2$, $3$, or $10$ (all values larger than $10$ are in the $10$ bin). 

Thus, using the uncertainty-aware prediction rule above, the classifier can use its uncertainty to adaptively return a longer or shorter list of predictions. This not only allows it to improve accuracy over a hard top-1 threshold. Arguably, the ability to vary the size of the predicted set of classes is a practically useful functionality in itself. As Figure \ref{fig:imagenet_betas} shows anecdotally, some of the labels (like ``notebook'' and ``laptop'') are semantically so similar to each other that it would seem only natural for the classifier to use them synonymously.

\begin{algorithm}[!t]
   \caption{Uncertainty-aware top-$k$}
   \label{alg:ua-top-k}
    \begin{algorithmic}
        \REQUIRE A Dirichlet parameter $\valpha \in \R^K$ obtained by applying the LB to the Gaussian over the logit of an input, a percentile threshold $T$ e.g. $0.05$, a function $\mathrm{class\_of}$ that returns the underlying class of a sorted index.
        \STATE
        \STATE $\tilde{\valpha} = \mathrm{sort\_descending}(\valpha)$ \COMMENT{start with the highest confidence}
        \STATE $\alpha_0 = \sum_i \alpha_i$
        \STATE $\mathcal{C} = \{ \mathrm{class\_of}(1) \}$ \COMMENT{initialize top-$k$, must include at least one class}
        
        \STATE $F_{1} = \mathrm{Beta}(\tilde{\alpha}_{1}, \alpha_0 - \tilde{\alpha}_{1})$ \COMMENT{the first marginal CDF}
        \STATE $l_{1} = F_{1}^\inv(T/2)$  \COMMENT{left $\frac{T}{2}$ percentile of the first marginal}

        \FOR{$i = 2, \dots, K$}
            \STATE $F_{i} = \mathrm{Beta}(\tilde{\alpha}_i, \alpha_0 - \tilde{\alpha}_i)$ \COMMENT{the current marginal CDF}
            \STATE $r_{i} = F_i^\inv(1-T/2)$ \COMMENT{right $\frac{T}{2}$ percentile of the current marginal}
            \IF{$r_i > l_{1}$}
                \STATE $\mathcal{C} = \mathcal{C} \cup \{ \mathrm{class\_of}(i) \}$  \COMMENT{overlap detected, add the current class}
            \ELSE
                \BREAK \COMMENT{No more overlap, end the algorithm}
            \ENDIF
        \ENDFOR
        \STATE
        \ENSURE $\mathcal{C}$ \COMMENT{return the resulting top-$k$ prediction}
    \end{algorithmic}
\end{algorithm}
\vspace{-1em}
%%%%%%%%%%%%%%%%%%%%%%%%
\section{Conclusion}
\label{sec:conclusion}
We have adapted a previously developed approximation scheme for new use in Bayesian Deep Learning. Given a Gaussian approximation to the weight-space posterior of a NN (which can be constructed by various means, including another Laplace approximation), and an input, the Laplace Bridge analytically maps the marginal Gaussian prediction on the logits onto a Dirichlet distribution over the softmax vectors. The associated computational cost of $\mathcal{O}(K)$ for $K$-class prediction compares favorably to that of MC sampling.
The proposed method empirically preserves predictive uncertainty, offering an attractive, low-cost, high-quality alternative to Monte Carlo sampling. In conjunction with a low-cost, last-layer Bayesian approximation, it is useful in real-time applications wherever uncertainty is required---especially because it drastically reduces the cost of predicting a posterior distribution at test time for a minimal increase in cost at training time.
The vanilla LB has some limitations, for which we proposed a simple correction that outperforms alternative softmax-integral approximations such as the commonly used multi-class probit.
We demonstrate the utility of the scheme for large-scale Bayesian inference by using it to construct an uncertainty-aware top-$k$ ranking on ImageNet.

\begin{contributions} % will be removed in pdf for initial submission,
                      % so you can already fill it to test with the
                      % ‘accepted’ class option
    MH wrote the code, created the figures, and wrote most of the paper. AK gave guidance and supervision and assisted throughout the entire process and greatly helped with the rebuttal. PH had the original idea and provided supervision. 
\end{contributions}

\begin{acknowledgements} % will be removed in pdf for initial submission,
                         % so you can already fill it to test with the
                         % ‘accepted’ class option
    The authors gratefully acknowledge financial support by the European Research Council through ERC StG Action 757275 / PANAMA; the DFG Cluster of Excellence “Machine Learning - New Perspectives for Science”, EXC 2064/1, project number 390727645; the German Federal Ministry of Education and Research (BMBF) through the Tübingen AI Center (FKZ: 01IS18039A); and funds from the Ministry of Science, Research and Arts of the State of Baden-Württemberg. 
    MH \& AK are grateful to Alexander Meinke
    for the pre-trained models and the International Max Planck
    Research School for Intelligent Systems (IMPRS-IS) for
    support. MH \& AK would also like to thank all members of Methods of Machine Learning group for helpful feedback.
\end{acknowledgements}

\bibliography{LB_for_BNNs}

\begin{thebibliography}{46}
\providecommand{\natexlab}[1]{#1}
\providecommand{\url}[1]{\texttt{#1}}
\expandafter\ifx\csname urlstyle\endcsname\relax
  \providecommand{\doi}[1]{doi: #1}\else
  \providecommand{\doi}{doi: \begingroup \urlstyle{rm}\Url}\fi

\bibitem[Ahmed and Xing(2007)]{ahmed2007tight}
Amr Ahmed and Eric Xing.
\newblock On tight approximate inference of the logistic-{N}ormal topic
  admixture model.
\newblock In \emph{Proceedings of the 11th Tenth International Workshop on
  Artificial Intelligence and Statistics}, 2007.

\bibitem[Begoli et~al.(2019)Begoli, Bhattacharya, and Kusnezov]{UIinMedicine}
E.~Begoli, T.~Bhattacharya, and D.~Kusnezov.
\newblock The need for uncertainty quantification in machine-assisted medical
  decision making.
\newblock \emph{Nat Mach Intell}, 1:\penalty0 20–23, 2019.

\bibitem[Blundell et~al.(2015)Blundell, Cornebise, Kavukcuoglu, and
  Wierstra]{Blundell2015WeightUI}
Charles Blundell, Julien Cornebise, Koray Kavukcuoglu, and Daan Wierstra.
\newblock Weight uncertainty in neural network.
\newblock In \emph{ICML}, pages 1613--1622. PMLR, 2015.

\bibitem[Braun and McAuliffe(2010)]{braun2010variational}
Michael Braun and Jon McAuliffe.
\newblock {V}ariational {i}nference for large-scale models of discrete choice.
\newblock \emph{Journal of the American Statistical Association}, 105\penalty0
  (489):\penalty0 324--335, 2010.

\bibitem[Brosse et~al.(2020)Brosse, Riquelme, Martin, Gelly, and
  Moulines]{brosse2020last}
Nicolas Brosse, Carlos Riquelme, Alice Martin, Sylvain Gelly, and {\'E}ric
  Moulines.
\newblock On last-layer algorithms for classification: Decoupling
  representation from uncertainty estimation.
\newblock \emph{arXiv preprint arXiv:2001.08049}, 2020.

\bibitem[Bulatov(2011)]{notMNIST2011}
Yaroslav Bulatov.
\newblock not{MNIST} dataset, 2011.
\newblock URL
  \url{http://yaroslavvb.blogspot.com/2011/09/notmnist-dataset.html}.

\bibitem[Clanuwat et~al.(2018)Clanuwat, Bober{-}Irizar, Kitamoto, Lamb,
  Yamamoto, and Ha]{KMNIST2018}
Tarin Clanuwat, Mikel Bober{-}Irizar, Asanobu Kitamoto, Alex Lamb, Kazuaki
  Yamamoto, and David Ha.
\newblock Deep learning for classical {J}apanese literature.
\newblock \emph{arXiv}, abs/1812.01718, 2018.

\bibitem[Dangel et~al.(2020)Dangel, Kunstner, and Hennig]{dangel2020backpack}
Felix Dangel, Frederik Kunstner, and Philipp Hennig.
\newblock Backpack: Packing more into backprop.
\newblock In \emph{International Conference on Learning Representations}, 2020.

\bibitem[Daxberger et~al.(2021)Daxberger, Kristiadi, Immer, Eschenhagen, Bauer,
  and Hennig]{laplace2021}
Erik Daxberger, Agustinus Kristiadi, Alexander Immer, Runa Eschenhagen,
  Matthias Bauer, and Philipp Hennig.
\newblock Laplace redux--effortless {B}ayesian deep learning.
\newblock In \emph{{N}eur{IPS}}, 2021.

\bibitem[Gibbs(1997)]{PhDGibbs}
Mark~N. Gibbs.
\newblock \emph{{B}ayesian {G}aussian Processes for Regression and
  Classification}.
\newblock PhD thesis, University of Cambridge, September 1997.

\bibitem[Graves(2011)]{Graves2011VB}
Alex Graves.
\newblock Practical {V}ariational {I}nference for neural networks.
\newblock In J.~Shawe-Taylor, R.~S. Zemel, P.~L. Bartlett, F.~Pereira, and
  K.~Q. Weinberger, editors, \emph{Advances in Neural Information Processing
  Systems 24}, pages 2348--2356. Curran Associates, Inc., 2011.

\bibitem[Haussmann et~al.(2019)Haussmann, Gerwinn, and
  Kandemir]{haussmann2019BEDLwithPAC}
Manuel Haussmann, Sebastian Gerwinn, and Melih Kandemir.
\newblock {B}ayesian evidential deep learning with pac regularization, 2019.

\bibitem[He et~al.(2016)He, Zhang, Ren, and Sun]{2015_ResNet}
Kaiming He, Xiangyu Zhang, Shaoqing Ren, and Jian Sun.
\newblock Deep residual learning for image recognition.
\newblock In \emph{Proceedings of the IEEE conference on computer vision and
  pattern recognition}, pages 770--778, 2016.

\bibitem[Hein et~al.(2019)Hein, Andriushchenko, and Bitterwolf]{Hein_2019_CVPR}
Matthias Hein, Maksym Andriushchenko, and Julian Bitterwolf.
\newblock Why relu networks yield high-confidence predictions far away from the
  training data and how to mitigate the problem.
\newblock In \emph{The IEEE Conference on Computer Vision and Pattern
  Recognition (CVPR)}, June 2019.

\bibitem[Hendrycks and Gimpel(2016)]{HendycksOODBaseline}
Dan Hendrycks and Kevin Gimpel.
\newblock A baseline for detecting misclassified and out-of-distribution
  examples in neural networks.
\newblock \emph{arXiv}, abs/1610.02136, 2016.

\bibitem[Hennig(2010)]{Hennig2010}
P.~Hennig.
\newblock \emph{Approximate Inference in Graphical Models}.
\newblock PhD thesis, University of Cambridge, November 2010.

\bibitem[Hennig et~al.(2012)Hennig, Stern, Herbrich, and
  Graepel]{KernelTopicModels2012}
P.~Hennig, D.~Stern, R.~Herbrich, and T.~Graepel.
\newblock Kernel topic models.
\newblock In \emph{Fifteenth International Conference on Artificial
  Intelligence and Statistics}, volume~22 of \emph{JMLR Proceedings}, pages
  511--519. JMLR.org, 2012.

\bibitem[Kristiadi et~al.(2020)Kristiadi, Hein, and Hennig]{kristiadi2020being}
Agustinus Kristiadi, Matthias Hein, and Philipp Hennig.
\newblock Being {B}ayesian, even just a bit, fixes overconfidence in relu
  networks.
\newblock In \emph{ICML}, pages 5436--5446. PMLR, 2020.

\bibitem[Krizhevsky et~al.(2014)Krizhevsky, Nair, and
  Hinton]{krizhevsky2014cifar}
Alex Krizhevsky, Vinod Nair, and Geoffrey Hinton.
\newblock The {CIFAR}-10 dataset.
\newblock \emph{online: http://www. cs. toronto. edu/kriz/cifar. html}, 55,
  2014.

\bibitem[LeCun(1998)]{MNISTLeCun}
Y.~LeCun.
\newblock The {MNIST} database of handwritten digits.
\newblock \emph{http://yann.lecun.com/exdb/mnist/}, 1998.

\bibitem[Louizos and Welling(2016)]{louizos_structured_2016}
Christos Louizos and Max Welling.
\newblock Structured and efficient {V}ariational deep learning with matrix
  {G}aussian posteriors.
\newblock In \emph{ICML}, 2016.

\bibitem[Lu et~al.(2020)Lu, Ie, and Sha]{LuProbit}
Zhiyun Lu, Eugene Ie, and Fei Sha.
\newblock Uncertainty estimation with infinitesimal jackknife, its distribution
  and mean-field approximation.
\newblock \emph{CoRR}, abs/2006.07584, 2020.
\newblock URL \url{https://arxiv.org/abs/2006.07584}.

\bibitem[MacKay(1992{\natexlab{a}})]{MacKay1992}
David J.~C. MacKay.
\newblock A practical {B}ayesian framework for backpropagation networks.
\newblock \emph{Neural Comput.}, 4\penalty0 (3):\penalty0 448–472, May
  1992{\natexlab{a}}.
\newblock ISSN 0899-7667.

\bibitem[Mackay(1995)]{McKay1995NetworkBayesReview}
David J~C Mackay.
\newblock Probable networks and plausible predictions — a review of practical
  {B}ayesian methods for supervised neural networks.
\newblock \emph{Network: Computation in Neural Systems}, 6\penalty0
  (3):\penalty0 469--505, 1995.

\bibitem[MacKay(1992{\natexlab{b}})]{mackay1992evidence}
David~JC MacKay.
\newblock The evidence framework applied to classification networks.
\newblock \emph{Neural computation}, 4\penalty0 (5):\penalty0 720--736,
  1992{\natexlab{b}}.

\bibitem[MacKay(1998)]{MacKay1998}
David~J.C. MacKay.
\newblock Choice of basis for laplace approximation.
\newblock \emph{Machine Learning}, 33\penalty0 (1):\penalty0 77--86, Oct 1998.
\newblock ISSN 1573-0565.

\bibitem[Malinin and Gales(2018)]{NIPS2018PriorNetworks}
Andrey Malinin and Mark Gales.
\newblock Predictive uncertainty estimation via prior networks.
\newblock In \emph{Advances in Neural Information Processing Systems}, pages
  7047--7058, 2018.

\bibitem[Malinin and Gales(2019)]{NIPS2019PriorNetworks_improved}
Andrey Malinin and Mark Gales.
\newblock Reverse {KL}-divergence training of prior networks: Improved
  uncertainty and adversarial robustness.
\newblock In \emph{Advances in Neural Information Processing Systems}, pages
  14520--14531, 2019.

\bibitem[Malinin et~al.(2019)Malinin, Mlodozeniec, and
  Gales]{malinin2019ensemble}
Andrey Malinin, Bruno Mlodozeniec, and Mark Gales.
\newblock Ensemble distribution distillation, 2019.

\bibitem[Martens and Grosse(2015)]{martens2015optimizing}
James Martens and Roger Grosse.
\newblock Optimizing neural networks with {K}ronecker-factored approximate
  curvature.
\newblock In \emph{ICML}, 2015.

\bibitem[McAllister et~al.(2017)McAllister, Gal, Kendall, van~der Wilk, Shah,
  Cipolla, and Weller]{McAllister2017ConcretePF}
Rowan McAllister, Yarin Gal, Alex Kendall, Mark van~der Wilk, Amar Shah,
  Roberto Cipolla, and Adrian Weller.
\newblock Concrete problems for autonomous vehicle safety: Advantages of
  {B}ayesian deep learning.
\newblock In \emph{IJCAI}, 2017.

\bibitem[Michelmore et~al.(2018)Michelmore, Kwiatkowska, and
  Gal]{AutoDrivingBayes}
Rhiannon Michelmore, Marta Kwiatkowska, and Yarin Gal.
\newblock Evaluating uncertainty quantification in end-to-end autonomous
  driving control.
\newblock \emph{CoRR}, abs/1811.06817, 2018.

\bibitem[Netzer et~al.(2011)Netzer, Wang, Coates, Bissacco, Wu, and
  Ng]{SVHN2011}
Yuval Netzer, Tao Wang, Adam Coates, Alessandro Bissacco, Bo~Wu, and Andrew~Y.
  Ng.
\newblock Reading digits in natural images with unsupervised feature learning.
\newblock In \emph{NIPS Workshop on Deep Learning and Unsupervised Feature
  Learning 2011}, 2011.

\bibitem[Nguyen et~al.(2015)Nguyen, Yosinski, and Clune]{nguyen2015deep}
Anh Nguyen, Jason Yosinski, and Jeff Clune.
\newblock Deep neural networks are easily fooled: High confidence predictions
  for unrecognizable images.
\newblock In \emph{CVPR}, 2015.

\bibitem[Pedregosa et~al.(2011)Pedregosa, Varoquaux, Gramfort, Michel, Thirion,
  Grisel, Blondel, Prettenhofer, Weiss, Dubourg, Vanderplas, Passos,
  Cournapeau, Brucher, Perrot, and Duchesnay]{scikit-learn}
F.~Pedregosa, G.~Varoquaux, A.~Gramfort, V.~Michel, B.~Thirion, O.~Grisel,
  M.~Blondel, P.~Prettenhofer, R.~Weiss, V.~Dubourg, J.~Vanderplas, A.~Passos,
  D.~Cournapeau, M.~Brucher, M.~Perrot, and E.~Duchesnay.
\newblock Scikit-learn: Machine learning in {P}ython.
\newblock \emph{Journal of Machine Learning Research}, 12:\penalty0 2825--2830,
  2011.

\bibitem[Ritter et~al.(2018)Ritter, Botev, and Barber]{ritter2018a}
Hippolyt Ritter, Aleksandar Botev, and David Barber.
\newblock A scalable laplace approximation for neural networks.
\newblock In \emph{International Conference on Learning Representations}, 2018.

\bibitem[Russakovsky et~al.(2014)Russakovsky, Deng, Su, Krause, Satheesh, Ma,
  Huang, Karpathy, Khosla, Bernstein, Berg, and Li]{ImageNet2015}
Olga Russakovsky, Jia Deng, Hao Su, Jonathan Krause, Sanjeev Satheesh, Sean Ma,
  Zhiheng Huang, Andrej Karpathy, Aditya Khosla, Michael~S. Bernstein,
  Alexander~C. Berg, and Fei{-}Fei Li.
\newblock Imagenet large scale visual recognition challenge.
\newblock \emph{CoRR}, abs/1409.0575, 2014.

\bibitem[Sensoy et~al.(2018)Sensoy, Kaplan, and Kandemir]{NIPS2018EvidentialDL}
Murat Sensoy, Lance Kaplan, and Melih Kandemir.
\newblock Evidential deep learning to quantify classification uncertainty.
\newblock In \emph{Advances in Neural Information Processing Systems}, pages
  3179--3189, 2018.

\bibitem[Snoek et~al.(2015)Snoek, Rippel, Swersky, Kiros, Satish, Sundaram,
  Patwary, Prabhat, and Adams]{ScalableBayesianOptimizationDNNs2015}
Jasper Snoek, Oren Rippel, Kevin Swersky, Ryan Kiros, Nadathur Satish,
  Narayanan Sundaram, Mostofa Patwary, Mr~Prabhat, and Ryan Adams.
\newblock Scalable {B}ayesian optimization using deep neural networks.
\newblock In Francis Bach and David Blei, editors, \emph{Proceedings of the
  32nd ICML}, volume~37 of \emph{Proceedings of Machine Learning Research},
  pages 2171--2180, Lille, France, 07--09 Jul 2015. PMLR.

\bibitem[Spiegelhalter and Lauritzen(1990)]{spiegelhalter1990sequential}
David~J Spiegelhalter and Steffen~L Lauritzen.
\newblock Sequential updating of conditional probabilities on directed
  graphical structures.
\newblock \emph{Networks}, 20\penalty0 (5):\penalty0 579--605, 1990.

\bibitem[Sun et~al.(2017)Sun, Chen, and Carin]{sun_learning_2017}
Shengyang Sun, Changyou Chen, and Lawrence Carin.
\newblock Learning structured weight uncertainty in {B}ayesian neural networks.
\newblock In \emph{Artificial {Intelligence} and {Statistics}}, pages
  1283--1292, 2017.

\bibitem[Titsias(2016)]{michalis2016one}
Michalis Titsias.
\newblock One-vs-each approximation to softmax for scalable estimation of
  probabilities.
\newblock In \emph{NIPS}, 2016.

\bibitem[Vadera et~al.(2020)Vadera, Jalaian, and Marlin]{vadera2020generalized}
Meet~P. Vadera, Brian Jalaian, and Benjamin~M. Marlin.
\newblock Generalized bayesian posterior expectation distillation for deep
  neural networks, 2020.

\bibitem[Wilson et~al.(2016)Wilson, Hu, Salakhutdinov, and
  Xing]{2016DeepKernelLearning}
Andrew~Gordon Wilson, Zhiting Hu, Ruslan Salakhutdinov, and Eric~P. Xing.
\newblock Deep kernel learning.
\newblock In Arthur Gretton and Christian~C. Robert, editors, \emph{Proceedings
  of the 19th International Conference on Artificial Intelligence and
  Statistics}, volume~51 of \emph{Proceedings of Machine Learning Research},
  pages 370--378, Cadiz, Spain, 09--11 May 2016. PMLR.

\bibitem[Wu et~al.(2018)Wu, Nowozin, Meeds, Turner, Hern{\'{a}}ndez{-}Lobato,
  and Gaunt]{DeterministicVI2018}
Anqi Wu, Sebastian Nowozin, Edward Meeds, Richard~E. Turner, Jos{\'{e}}~Miguel
  Hern{\'{a}}ndez{-}Lobato, and Alexander~L. Gaunt.
\newblock Fixing {V}ariational {B}ayes: Deterministic {V}ariational {I}nference
  for {B}ayesian neural networks.
\newblock \emph{arXiv}, abs/1810.03958, 2018.

\bibitem[Xiao et~al.(2017)Xiao, Rasul, and Vollgraf]{FMNIST2017}
Han Xiao, Kashif Rasul, and Roland Vollgraf.
\newblock Fashion-{MNIST}: a novel image dataset for benchmarking machine
  learning algorithms.
\newblock \emph{arXiv}, abs/1708.07747, 2017.

\end{thebibliography}

\newpage
\appendix
% NOTE: necessary when ptmx or no mathfont class option is given
\section{Appendix}
\label{appendix_A}
\subsection*{Figures}

The parameters of Figure \ref{fig:1D_Laplace_bridge} are from left to right $\alpha, \beta = (0.8,0.9), (4,2,), (2, 7)$.

\subsection*{Change of Variable for pdf} 
Let $\rvz$ be an $n$-dimensional continuous random variable with joint density function $p_\rvx$. If $\rvy = G(\rvx)$, where $G$ is a differentiable function, then $\rvy$ has density $p_\rvy$:
\begin{equation}
p(\mathbf{y}) = f\Big(G^{-1}(\mathbf{y})\Big)\left\vert \det\left[\frac{dG^{-1}(\mathbf{z})}{d\mathbf{z}}\Bigg \vert_{\mathbf{z}=\mathbf{y}}\right]\right \vert
\end{equation}
where the differential is the Jacobian of the inverse of $G$ evaluated at $\rvy$. This procedure, also known as `change of basis', is at the core of the Laplace bridge since it is used to transform the Dirichlet into the softmax basis.

\subsection*{Correction for sum(y)=0} 
We know that the product rule of Gaussians yields
\begin{align}
    p(x \vert Ax = y) &= \frac{p(x,y)}{p(y)} \\
    = \mathcal{N}(x; &\mu  + \Sigma A^\top (A \Sigma A^\top)^{-1} (y - A\mu),\\ \nonumber
    & \Sigma - \Sigma A^\top (A \Sigma A^\top)^{-1} A \Sigma )
\end{align}
In our particular setup we have
\begin{equation}
    p(x) = \mathcal{N}(x; \mu, \Sigma)
\end{equation}
with constraint
\begin{equation}
    p(I \vert x) = \delta(1 x^\top -0) = \lim_{\epsilon \rightarrow \infty} \mathcal{N}(0; 1^\top x, \frac{1}{\epsilon})
\end{equation}
Therefore we get
\begin{align}
    p(x \vert I) &= \mathcal{N}(x; \mu + \Sigma 1(1^\top \Sigma 1 - \frac{1}{\epsilon})^{-1}(0 - 1^\top \mu),\\ \nonumber
    &\Sigma - \Sigma 1(1^\top \Sigma 1 - \frac{1}{\epsilon})^{-1} 1^\top \Sigma) \\
    &= \mathcal{N}\left(x; \mu - \frac{\Sigma \mathbf{1} \mathbf{1}^\top \mu}{\mathbf{1}^\top \Sigma \mathbf{1}}, \Sigma - \frac{\Sigma \mathbf{1} \mathbf{1}^\top \Sigma}{\mathbf{1}^\top \Sigma \mathbf{1}} \right)
\end{align}

\subsection*{Variance correction} 
\begin{figure}
    \centering
    \includegraphics[width=0.48\textwidth]{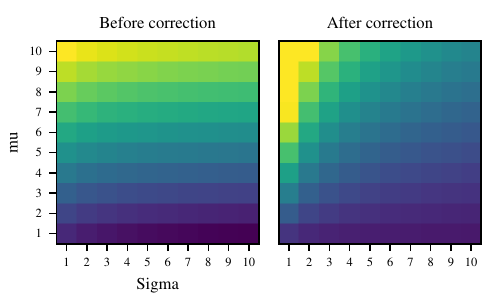}
    \caption{Contourplot showing the scaling behavior of $\mu$ and $\Sigma$. In the left figure, we see that Sigma has nearly no influence on the scaling. Our correction in the right figure fixes that. Contour levels show the first entry of $\alpha$ on a log-scale.}
    \label{fig:correction_contour}
\end{figure}

As described in the main text, the original Laplace Bridge scales worse with $\Sigma$ than sampling and applying the softmax. In Figure \ref{fig:correction_contour} you can see a contourplot that shows the scaling of mean and variance with and without correction. As suggested, the Variance has nearly no influence on the result before the correction but our correction fixes that. 

Some reviewers wanted to understand how we derived the equations for our correction, so here is a short informal explanation. During the experimentation with the LB, we found that it doesn't approximate the sample distribution well when $\Sigma$ gets large. We then understood why (as detailed in the limitations section) and proposed a fix for these scenarios without damaging its behavior in all other scenarios. We experimented with multiple fixes and the result you see in the paper is the one that fulfilled most of our criteria. Therefore, the correction doesn't come from a principled theoretical derivation but is motivated by the theoretical findings.

\section{Appendix (Derivation of LB)}
\label{appendix_B_LB}

%Notation
% vector p: axis in the standard basis == z
% vector a: axis in the transformed basis == \pi
% vector q: == \xi
% vector b: == \tau

Assume we have a Dirichlet in the standard basis with parameter vector $\valpha$ and probability density function:

\begin{equation}\label{eq:dirichlet_appendix}
    \mathrm{Dir}(\vpi | \valpha) := \frac{\Gamma \left( \sum_{k=1}^K \alpha_k \right)}{\prod_{k=1}^K \Gamma(\alpha_k)} \prod_{k=1}^K \pi_k^{\alpha_k-1} \, ,
\end{equation}

We aim to transform the basis of this distribution via the softmax transform to be in the new base $\pi$:

\begin{equation}\label{eq:softmax_appendix}
    \pi_k(\vz) := \frac{\exp(z_k)}{\sum_{l=1}^K \exp(z_l)} \, ,
\end{equation}

Usually, to transform the basis we would need the inverse transformation $H^{-1}(\vz)$ as described in the main paper. However, the softmax does not have an analytic inverse. Therefore David JC MacKay uses the following trick. Assume we know that the distribution in the transformed basis is:

\begin{equation}\label{eq:dirichlet_softmax_appendix}
    \mathrm{Dir}_{\vz}(\vpi(\vz) | \valpha) := \frac{\Gamma \left( \sum_{k=1}^K \alpha_k \right)}{\prod_{k=1}^K \Gamma(\alpha_k)} \prod_{k=1}^K \pi_k(\vz)^{\alpha_k} \, ,
\end{equation}

then we can show that the original distribution is the result of the basis transform by the softmax. 

\textbf{The Dirichlet in the softmax basis:} 
We show that the density over $\vpi$ shown in Equation \ref{eq:dirichlet_softmax_appendix} transforms into the Dirichlet over $\vz$. First, we consider the special case where $\vpi$ is confined to an $I-1$ dimensional subspace satisfying $\sum_i \vpi_i = c$. In this subspace we can represent $\varphi$ by an $I - 1$ dimensional vector $\varphi$ such that 

\begin{align}
    \pi_i &= \varphi_i \quad i,...,I-1 \\
    \pi_I &= c - \sum_i^{I-1} \varphi_i
\end{align}

and similarly we can represent $\vz$ by an $I-1$ dimensional vector $\va$:

\begin{align}
    z_i &= \va_i \quad i,...,I-1 \\
    z_I &= 1 - \sum_i^{I-1}\va_i
\end{align}

then we can find the density over $\va$ (which is proportional to the required density over $\vz$)
from the density over $\varphi$ (which is proportional to the given density over $\vpi$) by finding the
determinant of the $(I - 1) \times (I - 1)$ Jacobian $\mJ$ given by

\begin{align}
    J_{ik} &= \frac{\partial \varphi_i}{\partial \va_l} = \sum_j^{I} \frac{\partial \vpi_i}{\partial \rvz_j}\frac{\partial \rvz_j}{\partial \va_k} \nonumber\\
    &= \delta_{ik}\vpi_i - \vpi_i\vpi_k + \vpi_i\vpi_I =  \vpi_i(\delta_{ik} - (\vpi_k - \vpi_I))
\end{align}

We define two additional $I-1$ dimensional helper vectors $\rvz_k^+ := \rvz_k - \rvz_I$ and $n_k := 1$, and use $\det(I - xy^T) = 1 - x \cdot y$ from linear algebra. It follows that
\begin{align}
    \det J &= \prod_{i=1}^{I-1} \vpi_i \times \det[I-n\vpi^{+^T}] \nonumber \\
    &= \prod_{i=1}^{I-1} \vpi_i \times (1 - n \cdot \vpi^{+})  \\
    &= \prod_{i=1}^{I-1} \vpi_i \times \left(1 - \sum_k \vpi_k^{+} \right) = I \prod_{i=1}^I \vpi_i \nonumber
\end{align}

Therefore, using Equation \ref{eq:dirichlet_softmax_appendix} we find that
\begin{equation}
    P(\vpi) = \frac{P(\rvz)}{|\det \mJ|} \propto \prod_{i=1}^{I} \vpi_i^{\alpha_i - 1} 
\end{equation}
This result is true for any constant $c$ since it can be put into the normalizing constant. Thereby we make sure that the integral of the distribution is 1 and we have a valid probability distribution.

\section{Appendix (Derivation of Inversion)}
\label{appendix_C_inversion}
Through the figures of the 1D Dirichlet approximation in the main paper we have already established that the mode of the Dirichlet lies at the mean of the Gaussian distribution and therefore $\vpi(\vy) = \frac{\mathbf{\alpha}}{\sum_i \alpha_i}$. Additionally, the elements of $\vy$ must sum to zero. These two constraints combined yield only one possible solution for $\vmu$.

\begin{equation}
	\mu_k = \log \alpha_k  - \frac{1}{K} \sum_{l=1}^{K} \log \alpha_l
	\label{eq:mu_k}
\end{equation}

Calculating the covariance matrix $\vSigma$ is more complicated but layed out in the following. The logarithm of the Dirichlet is, up to additive constants

\begin{equation}
    \log p_\rvz(\rvz|\alpha) = \sum_k \alpha_k \pi_k 
\end{equation}

Using $\pi_k$ as the softmax of $\vy$ as shown in Equation \ref{eq:softmax_appendix} we can find the elements of the Hessian $\vL$

\begin{equation}
    L_{kl} = \hat{\alpha}(\delta_{kl}\hat{\pi_k} - \hat{\pi_k} \hat{\pi_l})
\end{equation}

where $\hat{\valpha} := \sum_k \alpha_k$ and $\hat{\vpi} = \frac{\alpha_k}{\hat{\alpha}}$ for the value
of $\vpi$ at the mode. Analytically inverting $\vL$ is done via a lengthy derivation using the fact that we can write $\vL = \mA + \mX\mB\mX^\top$ and inverting it with the Schur-complement. You can find the derivation in \citep{Hennig2010}. This process results in the inverse of the Hessian

\begin{equation}
    L_{kl}^{-1} = \delta_{kl} \frac{1}{\alpha_k} - \frac{1}{K} \left[\frac{1}{\alpha_k} + \frac{1}{\alpha_l} - \frac{1}{K}\left(\sum_u^K \frac{1}{\alpha_u}\right) \right]
\end{equation}

We are mostly interested in the diagonal elements, since we desire a sparse encoding for computational reasons and we otherwise needed to map a $K \times K$ covariance matrix to a $K\times 1$ Dirichlet parameter vector which would be a very overdetermined mapping. Note that $K$ is a scalar not a matrix. The diagonal elements of $\vSigma = \vL^{-1}$ can be calculated as

\begin{equation}
    \label{eq:Hessian_diag}
    \Sigma_{kk} = \frac{1}{\alpha_k} \left(1 - \frac{2}{K}\right)  + \frac{1}{K^2} \sum_{l}^{k} \frac{1}{\alpha_l}.
\end{equation}

To invert this mapping we transform Equation \ref{eq:mu_k} to 

\begin{equation}
    \label{eq:reform_mu_k}
    \alpha_k = e^{\mu_k} \prod_l^{K} \alpha_l^{1/K}
\end{equation}

by applying the logarithm and re-ordering some parts. Inserting this into Equation \ref{eq:Hessian_diag} and re-arranging yields

\begin{equation}
    \prod_l^K \alpha_l^{1/K} = \frac{1}{\vSigma_{kk}} \left[e^{-\mu}\left(1 - \frac{2}{K}\right)  + \frac{1}{K^2} \sum_u^K e^{-\mu_u} \right]
\end{equation}

which can be re-inserted into Equation \ref{eq:reform_mu_k} to give

\begin{equation}
    \label{eq:mapping_alpha_appendix}
    \alpha_k = \frac{1}{\Sigma_{kk}} \left(1 - \frac{2}{K} + \frac{e^{\mu_k}}{K^2} \sum_l^K e^{-\mu_k} \right)
\end{equation}

which is the final mapping. With Equations \ref{eq:mu_k} and \ref{eq:Hessian_diag} we are able to map from Dirichlet to Gaussian and with Equation \ref{eq:mapping_alpha_appendix} we are able to map the inverse direction.

\section{Appendix (Experimental Details)}
\label{appendix_D_experiments}
%%%%%%%%%%%%%%%%%%%%%
\begin{table*}[htb!]
    \scriptsize
    \fontsize{9}{10}\selectfont
    \setlength{\tabcolsep}{4pt}
    \centering
    \caption{Comparing the extended probit approximation with the normalized version of the LB norm in the KFAC setting. The probit approximation seems to break down in the MNIST scenarios. 
    }
    %#######################################
    %\resizebox{\textwidth}{!}{% use resizebox with textwidth
    \begin{tabular}{l  l || c c c c | c  c  c  c }
         \toprule
         & & \multicolumn{4}{c}{\textbf{KFAC Probit}} & \multicolumn{4}{c}{\textbf{KFAC LB norm}} \\
         \textbf{Train} & \textbf{Test} & \textbf{MMC} $\downarrow$ & \textbf{AUROC} $\uparrow$& \textbf{ECE} $\downarrow$ & \textbf{NLL} $\downarrow$& \textbf{MMC} $\downarrow$ & \textbf{AUROC} $\uparrow$& \textbf{ECE} $\downarrow$ & \textbf{NLL} $\downarrow$\\
         \midrule
          MNIST &    MNIST &            0.105 &              0.000 &            2.258 &            0.883 &             0.975 &               0.000 &             0.043 &             0.018 \\
  MNIST &   FMNIST &            0.102 &              0.955 &            2.302 &            0.032 &             0.444 &               0.990 &             2.871 &             0.364 \\
  MNIST & notMNIST &            0.103 &              0.922 &            2.300 &            0.043 &             0.409 &               0.986 &             2.854 &             0.294 \\
  MNIST &   KMNIST &            0.102 &              0.962 &            2.304 &            0.012 &             0.414 &               0.991 &             3.162 &             0.328 \\
  \midrule
CIFAR10 &  CIFAR10 &            0.548 &              0.000 &            0.661 &            0.404 &             0.941 &               0.000 &             0.195 &             0.017 \\
CIFAR10 & CIFAR100 &            0.358 &              0.896 &            2.652 &            0.253 &             0.662 &               0.866 &             3.871 &             0.558 \\
CIFAR10 &     SVHN &            0.307 &              0.956 &            2.567 &            0.195 &             0.441 &               0.965 &             2.837 &             0.327 \\ 
         \bottomrule
    \end{tabular}
    %}
    \label{tab:probit_table_kfac}
    %\vspace{-0.75em}
\end{table*}
%%%%%%%%%%%%%%%%%%%%%%%%

The exact experimental setups, i.e. network architectures, learning rates, random seeds, etc. can be found in the accompanying GitHub repository  
\footnote{\url{https://github.com/mariushobbhahn/LB_for_BNNs_official}}.
This section is used to justify some of the decisions we made during the process in more detail, highlight some miscellaneous interesting things and showcase the additional experiments promised in the main paper.

\subsection*{Mathematical description of the setup}

In principle, the Gaussian over the weights required by the Laplace Bridge for BNNs can be constructed by any Gaussian approximate Bayesian method such as variational Bayes \citep{Graves2011VB,Blundell2015WeightUI} and Laplace approximations for NNs \citep{MacKay1992,ritter2018a}. We will focus on the Laplace approximation, which uses the same principle as the Laplace Bridge. However, in the Laplace approximation for neural networks, the posterior distribution over the weights of a network is the one that is approximated as a Gaussian, instead of a Dirichlet distribution over the outputs as in the Laplace Bridge.

Given a dataset $\D := \{ (\vx_i, t_i) \}_{i=1}^D$ and a prior $p(\vtheta)$, let
\begin{equation}
    p(\vtheta | \D) \propto p(\vtheta) p(\D | \vtheta) = p(\vtheta) \prod_{(\vx, t) \in \D} p(y = t | \vtheta, \vx) \, ,
\end{equation}
be the posterior over the parameter $\vtheta$ of an $L$-layer network $f_\vtheta$. Then we can get an approximation of the posterior $p(\vtheta | \D)$ by fitting a Gaussian $\N(\vtheta | \vmu_\vtheta, \mSigma_\vtheta)$ where
\begin{align*}
    \vmu_\vtheta &= \vtheta_\text{MAP} \, , \\
    \mSigma_\vtheta &= (-\nabla^2 \vert_{\vtheta_\text{MAP}} \log p(\vtheta | \D))^\inv =: \mH_\vtheta^\inv \, .
\end{align*}
That is, we fit a Gaussian centered at the mode $\vtheta_\text{MAP}$ of $p(\vtheta | \D)$ with the covariance determined by the curvature at that point. We assume that the prior $p(\vtheta)$ is a zero-mean isotropic Gaussian $\N(\vtheta | \mathbf{0}, \sigma^2 \mI)$ and the likelihood function is the Categorical density
\begin{equation*}
    p(\D | \vtheta) = \prod_{(\vx, t) \in \D} \mathrm{Cat}(y = t | \mathrm{softmax}(f_\vtheta(\vx))) \, .
\end{equation*}
For various applications in Deep Learning, an approximation with full Hessian is often computationally too expensive. Indeed, for each input $\vx \in \R^N$, one has to do $K$ backward passes to compute the Jacobian $\mJ(\vx)$. Moreover, it requires an $\mathcal{O}(PK)$ storage which is also expensive since $P$ is often in the order of millions. A cheaper alternative is to fix all but the last layer of $f_\vtheta$ and only apply the Laplace approximation on $\mW_L$, the last layer's weight matrix. This scheme has been used successfully by \citet{ScalableBayesianOptimizationDNNs2015,2016DeepKernelLearning,brosse2020last}, etc. and has been shown theoretically that it can mitigate overconfidence problems in ReLU networks \citep{kristiadi2020being}. In this case, given the approximate last-layer posterior
\begin{equation}
    p(\mW^L | \D) \approx \N(\vec(\mW^L) | \vec(\mW^L_\text{MAP}), \mH_{\mW^L}^\inv) \, ,
\end{equation}
one can efficiently compute the distribution over the logits. That is, let $\vphi: \R^N \to \R^{Q}$ be the first $L-1$ layers of $f_\vtheta$, seen as a feature map. Then, for each $\vx \in \R^N$, the induced distribution over the logit $\mW^L \vphi(\vx) =: \vz$ is given by
\begin{equation}
    p(\vz | \vx) = \N(\vz | \mW^L_\text{MAP} \vphi(\vx), (\vphi(\vx)^\top \otimes \mI) \mH_{\mW^L}^\inv (\vphi(\vx) \otimes \mI)) \, ,
\end{equation}
where $\otimes$ denotes the Kronecker product.

An even more efficient last-layer approximation can be obtained using a Kronecker-factored matrix normal distribution \citep{louizos_structured_2016,sun_learning_2017,ritter2018a}. That is, we assume the posterior distribution to be
\begin{equation}
    p(\mW^L | \D) \approx \MN(\mW^L | \mW^L_\text{MAP}, \mU, \mV) \, ,
\end{equation}
where $\mU \in \R^{K \times K}$ and $\mV \in \R^{Q \times Q}$ are the Kronecker factorization of the inverse Hessian matrix $\mH_{\mW^L}^\inv$ \citep{martens2015optimizing} and $\MN$ denotes the Matrix Normal distribution. In this case, for any $\vx \in \R^N$, one can easily show that the distribution over logits is given by
\begin{equation}
    p(\vz | \vx) = \N(\vz | \mW^L_\text{MAP} \vphi(\vx), (\vphi(\vx)^\top \mV \vphi(\vx))\mU) \, ,
\end{equation}
which is easy to implement and computationally cheap. Finally, and even more efficient, is a last-layer approximation scheme with a diagonal Gaussian approximate posterior, i.e. the so-called mean-field approximation. In this case, we assume the posterior distribution to be
\begin{equation}
    p(\mW^L | \D) \approx \N(\vec(\mW^L) | \vec(\mW^L_\text{MAP}), \diag{\vsigma^2}) \, ,
\end{equation}
where $\vsigma^2$ is obtained via the diagonal of the Hessian of the log-posterior w.r.t. $\vec(\mW^L)$ at $\vec(\mW^L_\text{MAP})$.

\subsection*{OOD Detection}

The test scenarios are: A two-layer convolutional network trained on the MNIST dataset \citep{MNISTLeCun}. The OOD datasets for this case are FMNIST \citep{FMNIST2017}, notMNIST \citep{notMNIST2011}, and KMNIST \citep{KMNIST2018}. For larger datasets, i.e.~CIFAR-10 \citep{krizhevsky2014cifar}, SVHN \citep{SVHN2011}, and CIFAR-100 \citep{krizhevsky2014cifar}, we use a ResNet-18 network \citep{2015_ResNet}. In all scenarios, the networks are well-trained with $99\%$ test accuracy on MNIST, $95.4\%$ on CIFAR-10, $76.6\%$ on CIFAR-100, and $100\%$ on SVHN. For the sampling baseline, we use $100$ posterior samples.

All network have been trained with conventional setups, i.e. we use ADAM with learning rate $1e-3$ and weight decay $5e-4$ for the MNIST experiments and SGD with a cosine annealing scheduler starting at learning rate $0.1$ and momentum $0.9$ for the CIFAR and SVHN experiments. 

\subsection*{Probit vs LB}

The KFAC setting of the probit comparison can be found in Table \ref{tab:probit_table_kfac}. Especially in the MNIST scenario the probit approximation seems to break down since even in-dist detection is at chance level. The LB, on the other hand, yields reasonable results. 

\end{document}